\documentclass[lettersize,journal]{IEEEtran}
\usepackage{amsmath,amsfonts}
\usepackage{algorithmic}
\usepackage{algorithm}
\usepackage{array}
\usepackage[caption=false,font=normalsize,labelfont=sf,textfont=sf]{subfig}
\usepackage{textcomp}
\usepackage{stfloats}
\usepackage{url}
\usepackage{verbatim}
\usepackage{graphicx}
\usepackage{cite}
\usepackage{makecell}
\usepackage[T1]{fontenc}

\usepackage{xspace}
\usepackage{subcaption}
\usepackage{hyperref}
\usepackage{multirow}
\usepackage{color}
\usepackage{tabularx}
\usepackage{indentfirst} 
\begin{document}
\title{Flexible Physical Camouflage Generation\\ Based on a Differential Approach}

\author{Yang Li, Wenyi Tan, Tingrui Wang,  Xinkai Liang and Quan Pan
\thanks{\textit{(Corresponding author: Yang Li.)}}
\thanks{Yang Li, Wenyi Tan, Tingrui Wang, and Quan Pan are with School of Automation, Northwestern Polytechnical University, Xi'an 710072, China. Xinkai Liang is with Laboratory of Science and Technology on Complex System Control and Intelligent Agent Cooperation, Beijing, China. (e-mail: liyangnpu@nwpu.edu.cn, tanwenyi@mail.nwpu.edu.cn).}
}

\markboth{Journal of \LaTeX\ Class Files,~Vol.~14, No.~8, August~2021}%
{Shell \MakeLowercase{\textit{et al.}}: A Sample Article Using IEEEtran.cls for IEEE Journals}


\maketitle

\begin{abstract}
To achieve attacks on object detection in the physical world, many existing studies focus on adversarial camouflage generation techniques, which involve applying adversarial textures onto the surface of the target. However, the neural renderers used in previous works cannot realistically simulate the physical world, and most texture rendering processes are implemented on 2D images, without fully considering the influence of the structural information of 3D models on texture overlay. To address these challenges, we introduce a novel neural rendering approach specifically tailored for adversarial camouflage within a versatile 3D rendering framework. Our method, named Flexible Physical-camouflage Attack (FPA), goes beyond traditional techniques by faithfully simulating lighting conditions, material variations of the camouflage, the angle and distance between the target and the camera, ensuring a nuanced and realistic representation of textures on a 3D target. To achieve this, we employ a generative approach that learns adversarial patterns from a diffusion model. This involves combining specially designed adversarial loss, smoothness loss, non-printable score loss, and concealment constraint loss to ensure the adversarial, printability, and concealable nature of camouflage in the physical world. Furthermore, we showcase the effectiveness of the proposed camouflage in sticker mode, demonstrating its ability to cover the target without compromising adversarial information. Through empirical and physical experiments, FPA exhibits strong performance in terms of attack success rate and transferability. Additionally, the designed sticker-mode camouflage, coupled with a concealment constraint, adapts to the environment, yielding diverse styles of texture. Our findings highlight the versatility and efficacy of the FPA approach in adversarial camouflage applications.
\end{abstract}


\section{Introduction}
\IEEEPARstart{P}{revious} research has confirmed the vulnerability of neural networks, particularly in computer vision \cite{li2023few,zou2023object}, natural language processing\cite{li2021graph}, and deep reinforcement learning\cite{LI2023103259,LI2022108965}.
Most of the work involves attacks in the digital space, such as disrupting facial recognition\cite{zolfi2022adversarial,xiao2021improving}, image classification tasks\cite{naseer2021improving,mahmood2021robustness}, etc., through carefully designed perturbations, adversarial patches. Though the patch can be printed out to deploy the physical attack, environmental factors such as view angle, light, hue, etc., deeply affect the attack. 
Therefore, it is more difficult to conduct an adversarial attack on the model in the physical world as the complex physical constraints (camera pose, etc.).

Recently, the camouflage attack has sprung out by optimizing the adversarial pattern which can be attached to the entire body of the target (e.g., individuals, vehicles, etc.). Different from the patched attack, the camouflage attack needs to take care of an all-round attack and non-rigid structures on the target surface. Therefore, the two most critical aspects of a camouflage attack involve the method of texture generation and the rendering technology applied to the target surface.
According to existing works, there are primarily two methods for texture generation: One approach is based on optimizing texture images, achieving full vehicle camouflage through repeated texture overlays\cite{zhang2018camou,wu2020physical,suryanto2022dta}. However, the same texture presents different effects on various surfaces of the vehicle, making it challenging for such camouflage to effectively deceive target detectors when faced with varying distances and perspectives.
Another approach is to treat the entire area of the target surface as the optimization objective\cite{wang2021dual,wang2022fca}, where each part of the texture corresponds to a specific region on the target's curved surface. This method provides better robustness in coping with variations in viewpoints and occlusions. However, this approach increases the size of the texture, resulting in longer rendering times and making optimization more challenging.
Another challenge in camouflage attacks is the rendering of textures onto the target surface. Because 3D targets are non-planar, simulating the coverage status in real-world environments becomes more difficult. To address this issue, some works use neural rendering to map textures\cite{wang2022fca,zhou2024rauca}. However, these methods consider the impact of environmental factors on camouflage only after obtaining the 2D rendered image and cannot simulate the interference of physical factors during the rendering process. Another approach, such as DTA\cite{suryanto2022dta} and ACTIVE\cite{suryanto2023active}, is based on repetitive projection or triplanar mapping, mapping adversarial textures onto targets in 2D images. This is essentially a 2D-to-2D rendering process, which inevitably leads to inaccuracies in texture shapes when dealing with non-planar projections. The aforementioned methods, from texture rendering to the generation of final adversarial samples, are not purely 3D processes. They do not sufficiently utilize the structural information of 3D target models.
The final optimization yields adversarial textures covering the entire vehicle. However, due to limitations in the rendering methods, it's not possible to directly unwrap the adversarial textures on the vehicle's surface into an image.
In actual deployment, the texture can only be applied by cropping the rendered images from fixed angles, which ignores angular transformations on the vehicle's surface. This approach increases the risk of adversarial information loss and lacks flexibility.

In response to the aforementioned challenges, we are the first to propose a comprehensive, multi-view, differentiable adversarial camouflage generation framework from 2D textures to 3D targets, termed Flexible Physically-camouflage Attack (FPA). Diverging from traditional neural rendering approaches, we fashioned a renderer for adversarial camouflage by leveraging a differentiable mesh renderer framework. 
Within this framework, we harnessed modular differentiable rendering functions to simulate authentic lighting conditions and material variations, ensuring a nuanced representation of the 3D target's textures. This approach took into account variables such as lighting and material properties, contributing to a more realistic rendering outcome. Then, we learn the adversarial pattern from the diffusion model in a generative way and combine the designed adversarial loss and covert constraint loss to guarantee the adversarial camouflage's adversarial and covert nature in the physical world. Meanwhile, the camouflage in sticker mode can easily cover the target in the physical world without losing too much adversarial information.
The contribution of this work can be summarized as follows:
\begin{itemize}
    \item We established a rendering system rooted in a robust 3D framework, sidestepping the intricacies associated with training neural renderers and achieving heightened rendering efficiency. This framework can randomly set the viewpoint of the 3D target, ambient light intensity, and material properties of the camouflage during the rendering process. It generates adversarial camouflage in the form of UV maps, minimizing the loss of adversarial information during physical deployment and enhancing flexibility in real-world applications.
    \item We are the first to propose using diffusion models to generate adversarial textures for 3D targets. We learn adversarial patterns from diffusion models, and the learned adversarial patterns are more realistic. This approach enables us to expedite the optimization process when generating higher-resolution adversarial camouflage while ensuring clearer camouflage textures.
    \item We introduce novel confrontation loss and concealment constraint loss, which effectively enhance the aggressiveness and concealment of camouflage, facilitating adaptation to diverse environments.
\end{itemize}

\section{Related Works}
Given the prevalent utilization of adversarial samples, there has been a notable shift in research focus towards physical-world attacks, predominantly manifesting as adversarial patches and adversarial camouflage.
\subsection{Physical Adversarial Patch}
A physical adversarial patch is a physical object or pattern that, when strategically placed or affixed to an object in the real world, can cause misclassification by computer vision systems. These patches are designed using algorithms and optimization techniques to exploit vulnerabilities in the target system. They often incorporate features that, when perceived by the model, lead to incorrect predictions.
Early adversarial patches, were employed to attack image classification\cite{brown2017adversarial,eykholt2018robust,zhong2022shadows} and object detection\cite{song2018physical,lee2019physical,liu2018dpatch,thys2019fooling,chen2019shapeshifter,wang2020can,wu2020making,xu2020adversarial} models by being strategically placed on the target objects.
The Expectation Over Transformation (EOT) algorithm, as proposed by Athalye et al.\cite{athalye2018synthesizing}, takes a step further in generating robust adversarial samples. This algorithm simulates real-world environmental factors, including angles, brightness, distance, and various transformation parameters, during the training process, enhancing the resilience of the adversarial patches.
Lee et al.\cite{lee2019physical} successfully employed untargeted PGD\cite{madry2017towards} and the EOT\cite{athalye2018synthesizing} algorithm to generate adversarial patches that were capable of effectively attacking the YOLOv3\cite{redmon2018yolov3} detection algorithm in the physical world. To further enhance the robustness of adversarial patches, Xu et al.\cite{xu2020adversarial} introduced the Thin Plate Spline Mapping\cite{bookstein1989principal} technique to simulate non-rigid deformations on the surface of clothing, such as wrinkles, that occur in real-world environments.
Huang et al.\cite{huang2020universal} optimized adversarial patches by introducing a set of transformations to mimic deformations and attributes.
On the other hand, adversarial patches often feature strong visual appeal, which significantly impacts their concealability when applied in the physical world. 
Therefore, Tan et al.\cite{tan2021legitimate} conducted an analysis considering three aspects: color characteristics, edge features, and texture features. They utilized a projection function in conjunction to constrain the characteristics of the adversarial patch. Kung et al.\cite{hu2021naturalistic} employed Generative Adversarial Networks (GANs) to learn image manifolds from real-world images, thereby generating physical adversarial patches for use in object detectors.

Although adversarial patches have proven effective in attacking models, it is crucial to acknowledge their limitations in terms of distance and angles. In light of these constraints, our work primarily centers on the generation of adversarial camouflage.

\subsection{Physical Adversarial Camouflage}
Adversarial camouflage, when applied to the surface of a target object, demonstrates sustained attack efficacy across multiple viewpoints, in contrast to adversarial patches which are effective only from specific angles and on a limited portion of the target. To optimize the camouflage on the target object's surface, early research employed methods such as clone networks, genetic algorithms, and 3D modeling\cite{zhang2019camou,wu2020physical}. For instance, Hu et al.\cite{hu2022adversarial} introduced the Toroidal-Cropping-based Expandable Generative Attack (TCEGA) strategy, generating adversarial textures that can be repetitively applied for attacks from various viewpoints. However, these methods did not effectively integrate non-rigid deformations of the three-dimensional object surface into the training process of textures, leading to suboptimal utilization of the target's structural information, which is crucial for real-world attacks.
The emergence of neural rendering \cite{kato2018neural} has revolutionized adversarial camouflage by allowing mapping onto three-dimensional model surfaces, thereby recreating realistic effects in real-world scenes. Leveraging the differentiable nature of neural rendering, a considerable body of work \cite{wang2021dual,wang2022fca,duan2021dpa,suryanto2022dta} has employed white-box methods to optimize and obtain optimal adversarial textures.
However, relying on simple repetitive projection and full-coverage methods has demonstrated limited effectiveness in real-world applications. Suryanto et al.\cite{suryanto2023active} enhanced the process of camouflage mapping by employing tri-planar mapping \cite{nicholson2008gpu}, thereby improving the generality and robustness of adversarial camouflage. Additionally, Hu et al. \cite{hu2023physically} utilized Voronoi diagrams and the Gumbel-softmax method to optimize camouflage textures, enhancing the texture mapping process in conjunction with 3D Thin Plate Spline transformations \cite{bookstein1989principal,donato2002approximate,tang2019augmentation}. This comprehensive approach aims to address the limitations of previous methods, providing consistency and clarity in the development of effective adversarial camouflage strategies.
However, in this work, a more flexible and easily implemented camouflage is proposed.

\section{Methodology}
\label{sec:workflow}
\begin{figure*}[htp]
	\centering
	\includegraphics[scale=0.5]{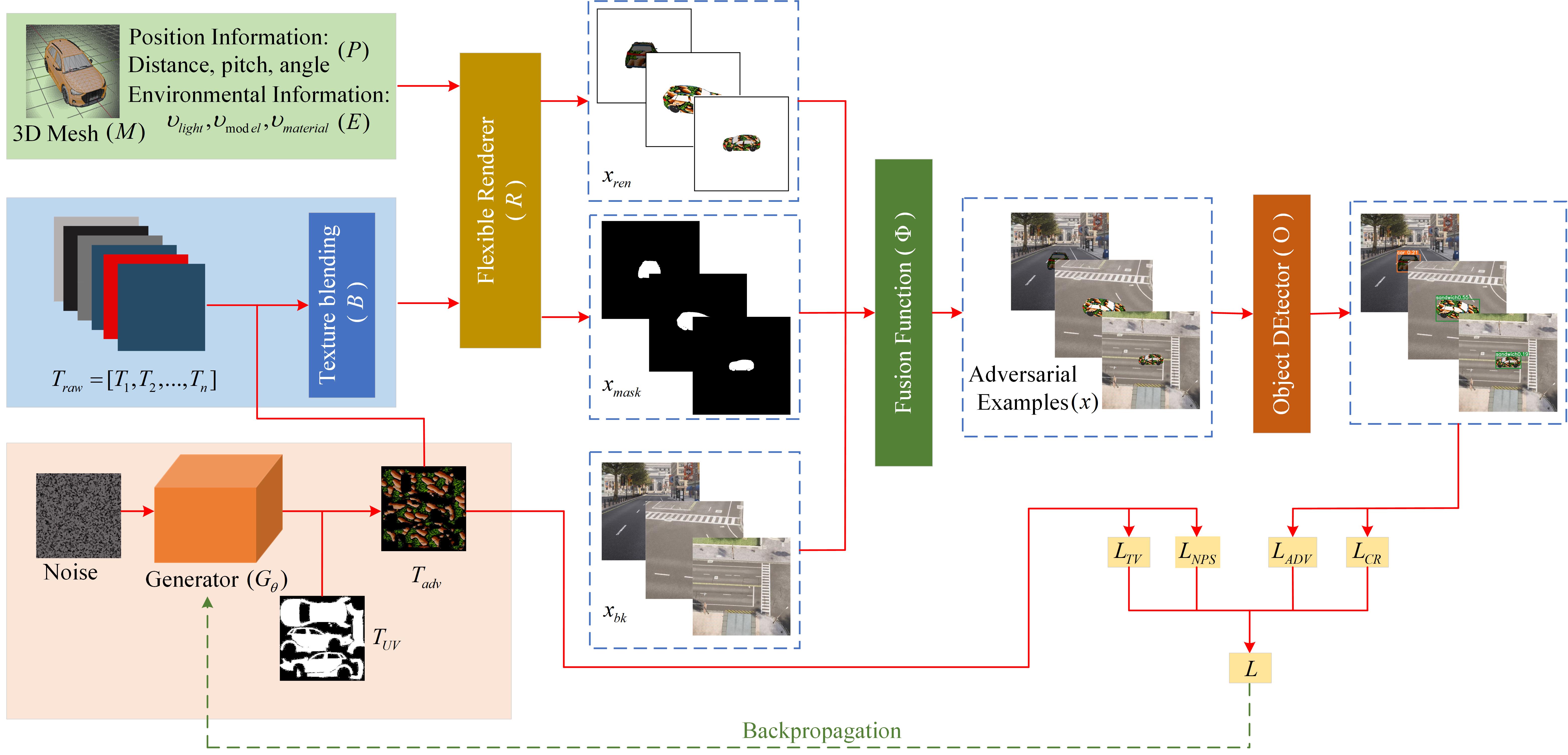}
	\caption{Workflow of generating robust adversarial textures in FPA framework.}
	\label{fig:fram}
\end{figure*}

In this section, we first briefly describe the process of generating adversarial camouflage for 3D objects, explaining the relationship between adversarial camouflage and the victim model, and outlining the requirements for successfully attacking detectors. Subsequently, we provide a detailed analysis of the various components of the attack framework FPA. The overall framework is as illustrated in Figure. \ref{fig:fram}.
\subsection{Problem Definition}
During the training process of adversarial camouflage, the attacker does not need the network structure and parameters of the target model, but only needs to provide input to the target model and obtain the predicted output. Unlike previous methods for training adversarial camouflage, we only need a small number of background images ${x_{bk}}$, without requiring a large dataset containing cars. Given texture \( T \), 3D mesh \( M \), and camera parameters \( P \) (which include the positional information between the object and the camera), differentiable renderer \( R \) can generate rendered images of the 3D object in different positions along with the corresponding mask images, $<x_{ren},x_{mask}>=R(T,M,P)$.
In $x_{mask}$, 0 represents the area where the object is located, and 1 represents the background area. By identifying the boundary points of the object area, we can obtain the true label \( y \) corresponding to $x_{ren}$.  We can merge the rendered image and the background using a fusion function $\Phi(\cdot)$, denoted as follows:
\begin{equation}
	\label{equ:mk}
 	\Phi(x_{ren},x_{mask},x_{bk})=x_{ren}\cdot x_{mask}+x_{bk}\cdot(1-x_{mask})
\end{equation}

Given a target detection model \( O \), upon inputting the target image, we can obtain the detector's predicted bounding box information $b=(b_{x},b_{y},b_{w},b_{h})$ (where $b_{x}$ and $b_{y}$ are the coordinates of the center of the predicted bounding box, and $b_{w}$ and $b_{h}$ represent the width and height of the predicted bounding box) and the confidence score $c$, collectively referred to as $<b^{(x)},c^{(x)}>={O}(x)$.

Our goal is to generate a robust adversarial texture that, when applied to the surface of a 3D object model through a differentiable renderer, can produce adversarial samples rendered from different positions. We formulate the generation of adversarial texture as an optimization problem, seeking the optimal adversarial texture that enables the object to evade detection by the target detector, as described below:
\begin{equation}
	\label{equ:pb}
	\underset{T_{\text {adv }}}{\arg \min } L\left({O}\left(\Phi\left(R\left(T_{a d v}, M, P\right), x\right)\right), y\right),
\end{equation}
where $L(\cdot)$ is the loss function.

\subsection{Camouflage Texture Generation}
Unlike the work of DTA\cite{suryanto2022dta} and ACTIVE\cite{suryanto2023active}, which repeatedly map optimized adversarial textures onto different surfaces of the target, we obtain the UV map of the 3D target surface texture through UV unwrapping. The UV map is a 2D texture map that can be accurately fitted on the 3D model. Subsequently, we can directly optimize the adversarial texture in the 2D space, i.e., the UV map $T_{UV}$ of the texture. Finally, the optimized adversarial texture is rendered onto the surface of the 3D target through the neural renderer. Therefore, our method achieves targeted optimization for all surfaces, ensuring that each region of the generated adversarial texture corresponds to a specific location on the 3D target surface. This guarantees the best attack effect from every viewpoint. As seen in the work of FCA\cite{wang2022fca} and RAUCA\cite{zhou2024rauca}, this approach makes the optimization of adversarial textures challenging, with the texture information being difficult to converge.

Since the emergence of diffusion models, their powerful generative capabilities have led to widespread application in the field of image generation\cite{ho2020denoising,song2020denoising,rombach2022high}. Therefore, we propose optimizing adversarial textures using a diffusion-based approach. We are the first to introduce the use of diffusion models to generate adversarial textures targeting 3D objects. Assuming the input is noise data $\epsilon$, typically following a Gaussian distribution $N(0, I)$, and the generator, based on U-Net\cite{ronneberger2015u} architecture, generates the adversarial texture $T_{k}$ at time $k$, each diffusion step can be described as follows: The generator takes the adversarial texture $T_{k}$ and adds Gaussian noise $\epsilon_{k}$ to it, where $\epsilon_{k}$ is a Gaussian distribution $N(0,\sigma_{k}^{2}I)$, and $\sigma_{k}$ is the standard deviation of the Gaussian noise at time $k$, which is shown as follows:
\begin{equation}
	\label{equ:dif}
	T_{k-1} = G_{\theta}(T_{k}+\epsilon_{k}),
\end{equation}
where $G_{\theta}$ is the generator with U-Net as the backbone, $\theta$ represents the parameters of the generator, and $k$ represents the time step.
The generator then applies a series of convolutional operations to the noisy image to obtain a new texture $T_{k-1}$. According to the shape of the UV map $T_{UV}$, obtain the desired adversarial texture  $T_{adv}$ from the generated texture $T_{k-1}$, i.e., $T_{adv}=T_{k-1}\cdot T_{UV}$.

\begin{figure}[!t]
	\centering
	\includegraphics[width=0.9\linewidth]{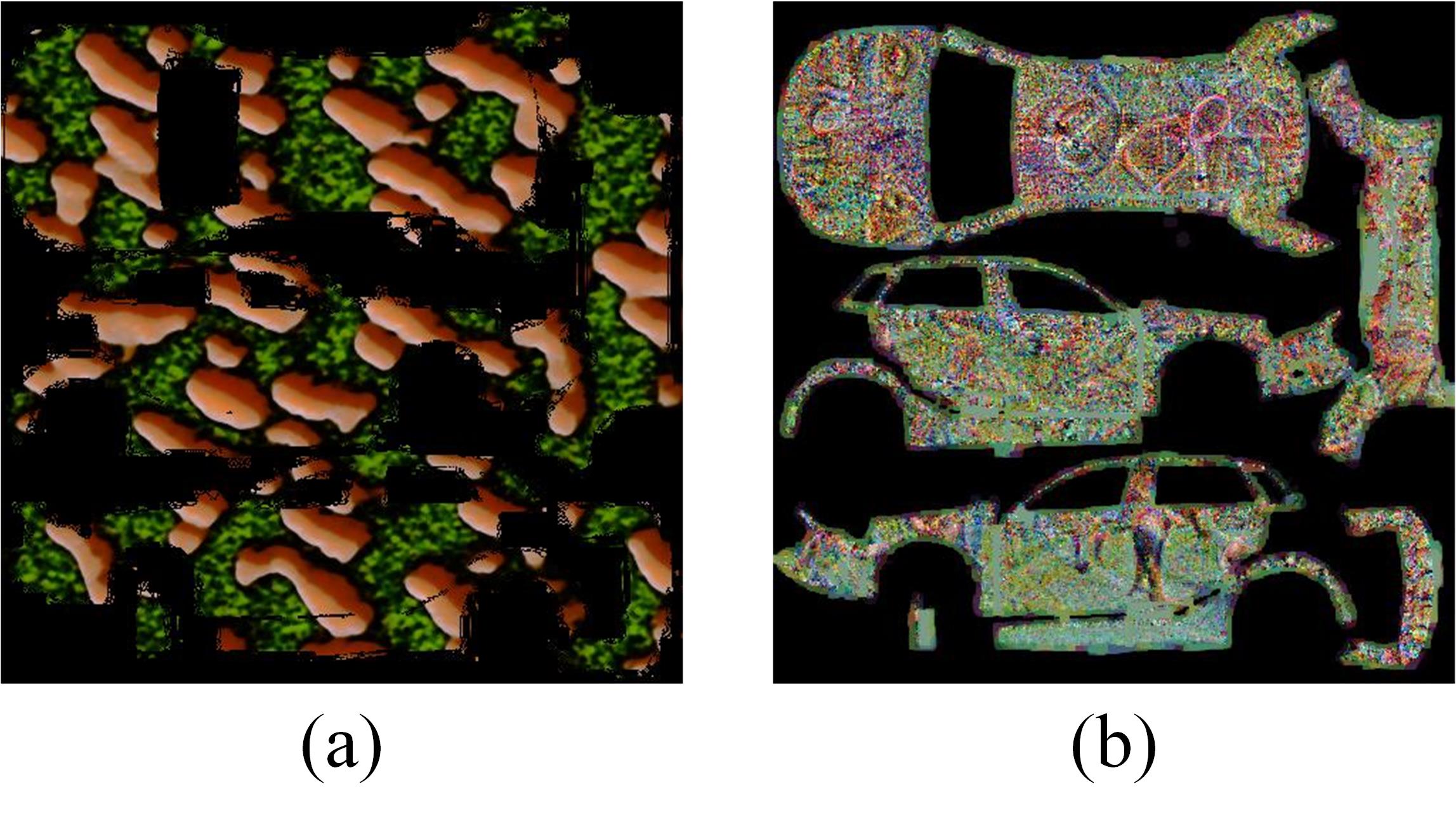}
	\caption{(a) shows the results generated based on the diffusion model, while (b) shows the results generated based on the gradient method.}
	\label{fig:duibi}
\end{figure}

Traditional gradient-based generation methods produce adversarial textures that are unclear and highly pixelated, failing to consider the impact of actual physical factors. We propose a method for learning adversarial patterns from diffusion models, which leverage diffusion models to capture various complex spatial relationships and interactions, ensuring the stable generation and clarity of adversarial textures. As shown in Figure. \ref{fig:duibi}, adversarial textures generated using diffusion models are more natural and legible . In the Section. \ref{sec:abl}, we further validate the generative capabilities of the diffusion model.

\subsection{Differential Rendering}
The difficulty of current adversarial camouflage generation is to map the 2D texture to the 3D object in a differential way. Additionally, it is difficult to transfer the generated adversarial texture into the physical world in an alignment manner. Therefore, we chose to use a neural renderer based on UV mapping to render the adversarial textures onto the surface of the 3D target. In works such as RACUA and FCA, the neural renderers\cite{kato2018neural} used independent light sources and were unable to render complex environmental features. Existing methods typically simulate environmental characteristics (e.g., EOT\cite{athalye2018synthesizing}, Automold\cite{automold}, EFE\cite{zhou2024rauca}) with the aid of external modules after obtaining the rendered images. This process is carried out on 2D images, ignoring the impact of the 3D structure (such as the occlusion of light by different parts of a car's surface or the reflectivity of the surface). To address these issues, we constructed the neural renderer based on a differentiable mesh\footnote{www.pytorch3d.org}, which supports the creation of new light sources during rendering and allows modifications to the surface material of the model. Then, in a differentiable manner, we render the adversarial texture onto the 3D target surface. This approach ensures that the adversarial texture is applied directly on the 3D model surface throughout the entire process, from rendering to the output of adversarial examples. 

{\textbf{Texture blending. }} The texture maps optimized through the diffusion model need to be rendered onto the corresponding regions of the 3D target for further optimization. One difficulty here is that almost all current 3D neural render methods can only process one texture, that is to say, there should only be one texture for a 3D model, otherwise, it can not be correctly rendered. To cope with this problem, the texture blending technique is applied to map different textures in a single 3D object.
The texture blending technique involves blending multiple textures based on the blending factor. The blending factor is the texture index associated with each vertex. The initial texture of the 3D model is $T_{raw}=[T_{1},T_{2},...,T_{n}]$. The blended texture can be represented as $B(T_{raw})=\sum_{i}^{n}\lambda_{i}\cdot T_{i}$, and $\lambda_{i}$ is the blending factor, which is typically set to 1. Therefore, we use a texture blending method to replace the corresponding regions in the original texture with the adversarial texture $T_{adv}$, without affecting the normal display of other areas on the 3D model surface. The final blended texture rendered onto the 3D model surface is described as follows:
\begin{equation}
	\label{equ:blend}
	T_{sum}=B(T_{raw},T_{adv})
\end{equation}

{\textbf{Flexible Renderer. }} Before generating the rendered images, we need to set the position information $P$ between the target and the camera, as well as the environmental information $E$ (including lighting intensity, surface material reflection parameters, etc.). The position information $P$ includes three parameters: distance, pitch angle, and azimuth angle, ensuring the generation of rendered images from different distances and viewpoints. Unlike previous work, the renderer $R$ built with PyTorch3D allows for the addition of new light sources in the rendering space. We set the parameter $\upsilon_{light}$ to control the light intensity and the parameter $\upsilon_{model}$ to control the lighting mode during the rendering process, including global illumination, directional light, and point light sources. In actual deployment, the camouflage's different material properties can lead to significant differences in texture appearance. We set the parameter $\upsilon_{material}$ to adjust the material of the camouflage during rendering, primarily controlling its reflectivity to light. Therefore, given the input 3D mesh $M$ and the blended texture $T_{sum}$, we can generate rendered images and corresponding mask images through the renderer, as shown in the following equation: 
\begin{equation}
	\label{equ:renderer}
	<x_{ren},x_{mask}>=R(T_{sum},M,P,E)
\end{equation}

Subsequently, applying the blending step combines the background and the rendered object in the same image to generate more diverse adversarial examples. Combining this process with the Equation~\ref{equ:mk}, the process can be described as follows:
\begin{equation}
	\label{equ:hunhe}
	x_{adv}=\phi(x_{ren},x_{mask},x_{bk})
\end{equation}

\begin{figure}[!t]
	\centering
	\includegraphics[width=1\linewidth]{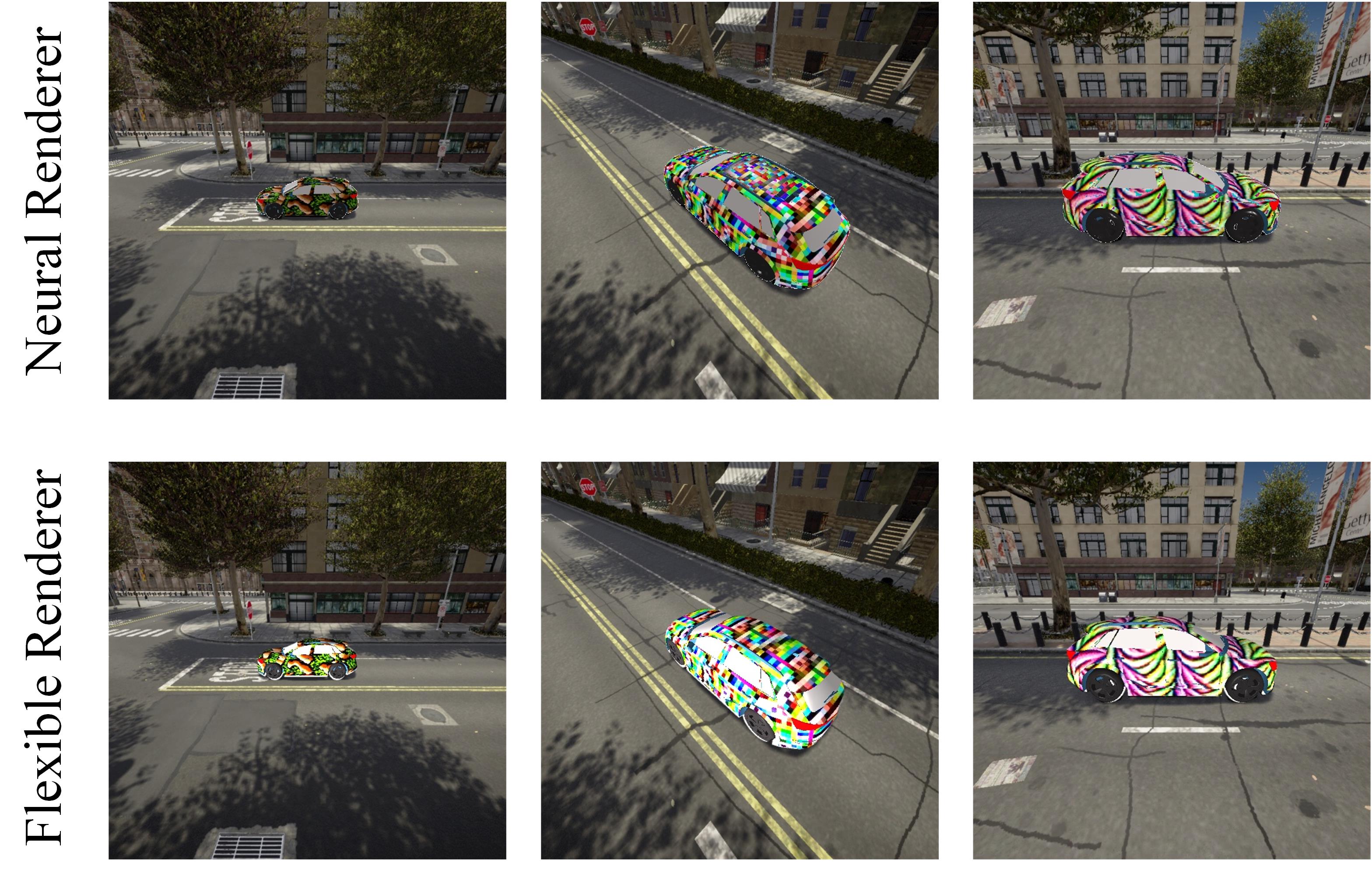}
	\caption{Comparison of rendered images with different textures using the Neural Renderer and the Flexible Renderer.}
	\label{fig:xuanran}
\end{figure}

As shown in Figure. \ref{fig:xuanran}, in scenes with direct lighting, compared to commonly used neural renderers, our method considers environmental information during the rendering process and employs UV mapping. This enables us to simulate more realistic environments and achieve more accurate texture projection.

The entire rendering process is detailed in Algorithm~\ref{alg:pytorch3d}. 
\begin{algorithm}
	\caption{Differential Rendering}
	\label{alg:pytorch3d}
	\begin{algorithmic}[1]
		\STATE {Set the batch size of the render images}
		\STATE {Load 3D object file $M$ with ``load objects as meshes''}
		\STATE {Load the initial texture $T_{raw}=[T_{1},T_{2},...,T_{n}]$ and adversarial texture $T_{adv}$ generated by the diffusion model} 
		\STATE {Obtain the blended texture $T_{sum}$ with Equation~\ref{equ:blend}}
		\STATE Create a Phong renderer by composing a rasterizer and a shader
		\FOR{i in batch size}
		\STATE Sample the elevation, azimuth, and distance from the defined range, denoted as parameters $P$
		\STATE Set the values of $\upsilon_{light}$, $\upsilon_{model}$, $\upsilon_{material}$, denoted as parameters $E$
		\STATE {Obtain the render images and mask images with Equation~\ref{equ:renderer}}
		\ENDFOR
	\end{algorithmic}
\end{algorithm}

\subsection{Loss constraints}
{\textbf{Adversarial Loss. }} After the object detector identifies the target, it outputs predicted confidence scores and the bounding box positions. Our objective is to evade detection by the detector, thus the designed adversarial loss ${L_{adv}}$ consists of two components: ${L_{iou}}$ and ${L_{obj}}$. Firstly, ${L_{iou}}$ is defined as the intersection over union (IOU) between the bounding boxes output by the object detector and the bounding boxes in the ground truth labels. During the non-maximum suppression process of object detection, bounding boxes with an IOU lower than a threshold are directly filtered out. Therefore, our objective is to reduce the overlap area of bounding boxes, thereby decreasing the IOU value.
${L_{obj}}$ is defined as the confidence score of the bounding boxes containing the target. We identify the bounding boxes with the highest confidence score as the target and optimize towards minimizing it during the training process, thereby suppressing the recognition of the object detector. Therefore, our adversarial loss is defined as follows:
\begin{equation}
	\label{equ:adv}
	\begin{cases}
		L_{iou}=max(IOU(y^{(x)},b_{i}^{(x)})),\\L_{obj}=max(c^{(x)}),\\	L_{ADV}=min(\alpha L_{iou}+\beta L_{obj})
	\end{cases},i=1,2,\ldots N,
\end{equation}
in the equation, ${x}$ represents the input image, ${y}$ denotes the ground truth label, ${N}$ is the number of output bounding boxes, ${c}$ stands for the confidence score predicted by the detector, ${b}$ represents the bounding boxes information predicted by the detector, and $\alpha$, $\beta$ are the weight coefficient.

{\textbf{Texture Smoothness. }} Texture smoothness is a key procedure to ensure the generated texture can be applied in the physical world with naturalness, effectiveness, and printability. As a smooth texture is more natural and realistic, making it less suspicious to human observers and more difficult to detect by deep learning systems, also improves the effectiveness of the adversarial attack by reducing the visual artifacts and noise that may interfere with the target object recognition, and is more printable and transferable to physical objects, making it easier to apply the adversarial attack in the real world.
Therefore, we propose the total variation loss in Equation~\ref{equ:tv}.
\begin{equation}
	\label{equ:tv}
	L_{TV} = \frac{1}{N_{nz}}\sum_{i,j}(T_{i,j}-T_{i+1,j})^2+(T_{i,j}-T_{i,j+1})^2,
\end{equation}
where $T_{i,j}$ is the pixel value in the texture $T$. Different from the original TV loss, each value is decided by the mask which is whether there is a texture pattern or not, and $N_{nz}$ is the total number of pixels in the area we selected. 

{\textbf{Digital to Physical Score. }} It is necessary to convert the digital color to the physical world to ensure the generated texture can be applied in the physical world. 
The generally used method is color management to ensure the color in digital and printable spaces is the same.
Let $P$ be the set of printable colors, and let $C$ be the set of colors in the adversarial texture. Let $d(T_i, p_j)$ be the Euclidean distance between color $T_i$ in the texture and color $p_j$ in the set of printable colors. Then, the non-print-ability score loss $L_{NPS}$ is given by:
\begin{equation}
	\label{equ:nps}
	L_{NPS}=\sum_{i=1}^{H} \sum_{j=1}^{W}\min_{j=1}^{|P|}d(T_{i,j},p_{j}),
\end{equation}
where $H$ and $W$ are the height and width of the adversarial texture, respectively. The goal of the optimizer is to minimize the non-printability score loss, along with other loss functions such as the total variation loss and objectness score loss, to generate an adversarial texture that can be physically printed and placed on an object to fool an object detection system.

{\textbf{Concealment Constraint. }} We propose a concealment constraint loss to ensure the similarity between the generated texture and the background, enhancing the dual concealment of texture in both machine vision and human visual observation. First, we compute the average RGB value ${c_{r}}$ within the background region based on the target. Based on the set color fluctuation threshold \( c_{r_u} \), the range of background color variation can be obtained as $[u_l, u_h]$, where \( u_l = c_r - c_{r_u} \) and \( u_h = c_r + c_{r_u} \). Then, we compute the difference between the generated texture and the background color variation to construct the concealment constraint loss, which can be represented as Equation~\ref{equ:color}.
\begin{equation}
	\label{equ:color}
	L_{CR} = \frac{1}{N}\sum_{i,j}(T_{i,j}-u_{l})^2+(T_{i,j}-u_{h})^2,
\end{equation}
where $N$ denotes the total number of the nonzero pixels, \(T_{i,j} \) represents the pixel value of adversarial texture at coordinate \( (i,j) \), lower \( L_{CR} \) indicates closer resemblance between the texture color and the background. Thus, minimizing \( L_{CR} \) enhances the similarity between texture color and background, thereby improving concealment in the environment.

The total loss of adversarial training is described by Equation~\ref{equ:L}, ${\gamma}, {\mu}, {\tau}$ are the weights to balance loss.
\begin{equation}
	\label{equ:L}
	L = L_{ADV} + {\gamma}L_{TV} + {\mu}L_{NPS} + {\tau}L_{CR}
\end{equation}

The training procedure is described in Algorithm~\ref{alg:2}.

\begin{algorithm}
    \caption{Image optimization with diffusion model}
    \label{alg:2}
    \begin{algorithmic}[1]
        \STATE {Load object detection model as the target model $\mathbf{O}$}
        \STATE {Load the background images $x_{bk}$ and the UV map $T_{UV}$}
        \FOR {step in STEPS}
        \FOR{k in Diffusion STEPS}
        \STATE {Obtain the adversarial texture $T_{k-1}$ with Equation~\ref{equ:dif}}
        \ENDFOR
        \STATE {Obtain the desired adversarial texture $T_{adv}$ from the generated texture $T_{k-1}$ with $T_{adv}=T_{k-1}\cdot T_{UV}$.}
        \STATE {Obtain the rendered image $x_{ren}$ and the mask image $x_{mask}$ with Algorithm~\ref{alg:pytorch3d}.}
        \STATE {Obtain the true label $y$ corresponding to $x_{ren}$ by identifying the boundaries of the target region in $x_{mask}$.}
        \STATE {Obtain the adversarial example $x$ with Equation~\ref{equ:hunhe}.}
        \STATE {<$b^{(x)}, c^{(x)}> = \mathbf{O}(x, y^{(x)})$}\
        \STATE {Obtain the adversarial loss with Equation~\ref{equ:adv}.}
        \STATE {Obtain the smooth variants with Equation~\ref{equ:tv}.}
        \STATE {Obtain the color constraints with Equation~\ref{equ:nps}.}
        \STATE {Obtain the concealment constraints with Equation~\ref{equ:color}.}
        \STATE {Set $L$ by Equation~\ref{equ:L}.}
        \STATE {Update diffusion model parameters $\theta$ for minimizing $L$ by via backpropagation.}
        \ENDFOR
    \end{algorithmic}
\end{algorithm}

\section{Experiments}
\label{sec:exp}
In this section, we first introduce the foundational configuration of the experiments, including experimental parameters, datasets, object detection algorithms, evaluation metrics, and more. Then, combining simulations and real-world environments, we conduct a thorough and adequate assessment of camouflage.
\subsection{Experimental Settings}
\noindent{\textbf{Datasets }}  To ensure the effectiveness of physical world attacks, we conducted experiments using a dataset collected based on the CARLA simulator\cite{dosovitskiy2017carla}. To facilitate fair comparisons with prior adversarial camouflage works, we directly employed the same dataset as (FCA\cite{wang2022fca}). The dataset was gathered in modern urban environments provided by CARLA, encompassing various distances, pitch angles, and yaw angles. The training set comprises 12,500 images, while the validation set consists of 3,000 images. Furthermore, to validate the superiority of our framework and loss function in camouflage concealment, we conducted experiments based on the Place365 dataset, selecting three common scenarios (grassland, desert, and highway). Each scene's training set comprises 50 background images, and the validation set comprises 10 images.

\noindent{\textbf{Target Models }}  To evaluate the adversarial camouflage's attack performance, we selected commonly used detection models of different architectures for experimentation. Specifically, these include single-stage detection algorithms such as SSD\cite{liu2016ssd}, YOLOv3\cite{redmon2018yolov3}, YOLOv5\cite{jocher10ultralytics}, and YOLOv7\cite{wang2023yolov7}, as well as two-stage detection algorithms like Faster R-CNN(FRCNN)\cite{ren2015faster}, Mask R-CNN(MkRCNN)\cite{he2017mask}, and Cascade R-CNN(CaRCNN)\cite{cai2018cascade}. Additionally, we included detection algorithms based on Transformer architecture, namely DETR\cite{carion2020end} and RT-DETR\cite{lv2023detrs}. Among these, SSD and Faster R-CNN were pretrained on the VOC2007 dataset\cite{everingham2010pascal}, while the remaining detection models were pretrained on the COCO dataset\cite{lin2014microsoft}. 

\noindent{\textbf{Evaluation Metrics }}  For object detection models, AP@0.5 is commonly used to evaluate the recognition capability of the models. We set the value of AP@0.5 at an IoU threshold of 0.5 as the first evaluation metric for assessing attack performance. The purpose of adversarial camouflage is to conceal the target vehicle and evade detection by the object detector. Therefore, we collect all candidate boxes detected by the detector for each image, and if the confidence scores of all candidate boxes are below the threshold conf=0.5, the image is considered successfully attacked. The second evaluation metric is the Attack Success Rate (ASR), defined as the proportion of successfully attacked images among all test images.

\noindent{\textbf{Implementation Details }}  In the actual deployment of adversarial camouflage, there is a process of magnification and printing. To minimize the loss of details in adversarial camouflage, considering the existing hardware conditions, we set a higher resolution for the adversarial camouflage, with an initial size of 480*480. The adversarial camouflage uses the same random seed for each training iteration, initialized with random noise. We employ Adam as the optimizer with default parameters, a learning rate of 0.001, and a maximum of 10 epochs. We set $\upsilon_{model}$ to define the lighting mode: in the range [0, 0.33] for point light mode, in the range (0.33, 0.67] for directional light mode, and in the range (0.67, 1] for global illumination mode. The range for the light intensity parameter $\upsilon_{light}$ is set to [0.2, 5], and the range for the material reflectivity $\upsilon_{material}$ is set to [0.5, 1.5]. The parameters of the loss function are configured based on extensive experimental experience, with the following settings: $\alpha=0.05$, $\beta=1.0$, $\gamma=1.0$, $\mu=2.5$, $\tau=2.0$. All our code is implemented in PyTorch and training/validation are conducted on an NVIDIA GeForce RTX 3090 GPU.

\subsection{Universality Evaluations}
\label{sec:4.2}
\noindent{\textbf{Transferability Evaluation Across Different Victim Models }} We trained adversarial camouflages against different victim models: FPA(v3) against YOLOv3, FPA(v5) against YOLOv5, and FPA(FR) against FRCNN.To ensure a fair comparison with previous adversarial camouflage works, we validated our approach using the same dataset as FCA. We added eight groups of adversarial camouflages as comparative experiments: Noise, CAMOU\cite{zhang2018camou}, ER\cite{wu2020physical}, DAS\cite{wang2021dual}, FCA\cite{wang2022fca}, DTA\cite{suryanto2022dta}, ACTIVE\cite{suryanto2023active}, RAUCA\cite{zhou2024rauca}. Among these, the adversarial camouflage CAMOU and ER cover the entire vehicle by repeated pasting. For DAS, FCA and RAUCA adversarial camouflage, we employed the same neural rendering technique as the original paper, overlaying textures onto the vehicle surface through neural rendering. For DTA and ACTIVE, we applied the repeated texture projection method to map them onto the target surface. To validate the effectiveness of the FPA framework, we re-generated adversarial textures FPA(bk) for YOLOv5 using 20 background images collected from Carla (excluding cars). 

\begin{figure*}[htp]
	\centering
 \includegraphics[scale=0.15]{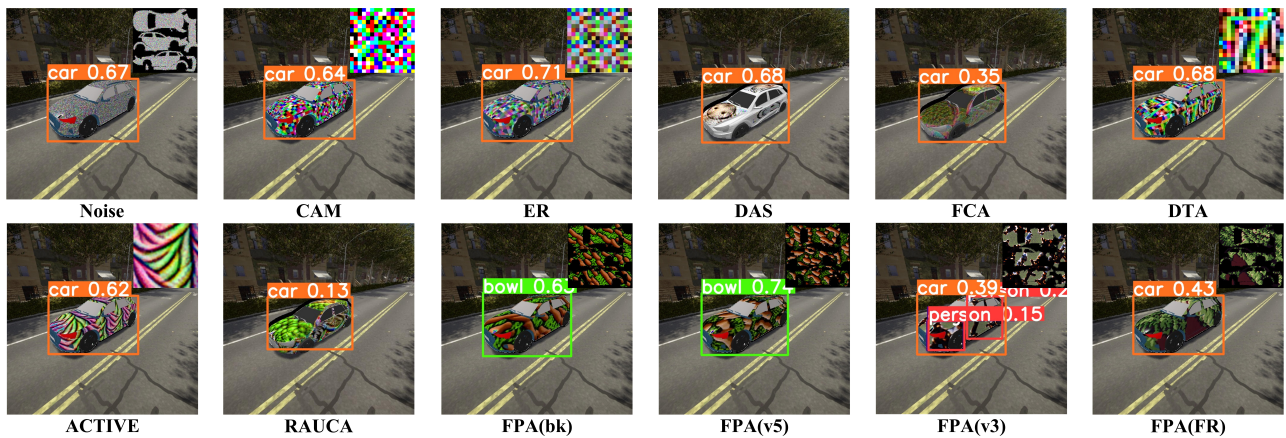}
	\caption{The presentation of adversarial camouflage and the prediction results under the YOLOv5 detection model.}
	\label{fig:2}
\end{figure*}

As shown in Figure.\ref{fig:2}, vehicles covered by our trained adversarial camouflage interfere with detector recognition and demonstrate the most effective attack. Table.\ref{lab:2} presents the comparison results of adversarial camouflage on AP@0.5. Our method achieves the best results on multiple detection models. Specifically, on YOLOv3, FPA(v5) exhibits the maximum drop of 73.9\% in AP@0.5, surpassing FCA by 10\%. On the widely used YOLOv5, it achieves an 89.1\% reduction, indicating the concealment capability of our camouflage at any angle. Due to the architectural differences of FRCNN and the diversity of training datasets for detection models, the effectiveness of FPA (v5) in attacks slightly decreased, yet it still achieved a 52\% reduction, surpassing other real-world attack methods. When employing FRCNN as the victim model, our adversarial camouflage, FPA(FR), achieved a reduction of 74.9\%. The last column in the table indicates the average decrease in AP@0.5 across the three detectors, with our method achieving the best results. In particular, FPA (v5) achieved a reduction of 71.6\%, demonstrating the ability of our framework to generate highly aggressive camouflage when faced with different victim detection models. The validation results of FPA(bk) are only slightly lower than those of FPA(v5), yet it still demonstrates strong attack capabilities. This shows that our framework can generate adversarial textures from a small number of background images, and the adversarial textures exhibit strong robustness against different detectors and viewpoint changes.

\begin{table}[!ht]
	\caption{The comparison results of AP@0.5 for adversarial camouflage generated based on different victim models.The "Drop\textcolor{red}{$\downarrow$}" column indicates the average decrease in AP@0.5 of the adversarial camouflage across the three detectors.}
	\centering
	\begin{tabular}{c|c|c|c|c}
		\hline
		Method & YOLOv3  & YOLOv5  & FRCNN & Drop\textcolor{red}{$\downarrow$} \\ \hline
		RAW & 0.898 & 0.979 & 0.839 & 0.000  \\ \hline
		Noise & 0.672 & 0.781 & 0.622 & 0.214 \\ \hline
		CAMOU & 0.660 & 0.654 & 0.581 & 0.274 \\ \hline
		ER & 0.743 & 0.804 & 0.734 &  0.145 \\ \hline
		DAS & 0.803 & 0.916 & 0.833 & 0.054 \\ \hline
		FCA & 0.265 & 0.563 & 0.545 &0.448 \\ \hline
		DTA & 0.546 & 0.517 & 0.637 & 0.339 \\ \hline
		ACTIVE & 0.497 & 0.438 & 0.526 & 0.418 \\ \hline
		RAUCA & 0.209 & 0.148 & 0.441 & 0.639 \\ \hline
		FPA(bk) & 0.146 & 0.104 & 0.348 & 0.706 \\ \hline
  		FPA(v3) & 0.212 & 0.343 & 0.487 & 0.558 \\ \hline
		FPA(v5) & \textbf{0.159} & \textbf{0.088} & \textbf{0.319} & \textbf{0.716} \\ \hline
		FPA(FR) & 0.543 & 0.513 & \textbf{0.090} & 0.523\\ \hline
	\end{tabular}
	\label{lab:2}
\end{table}

\begin{figure*}[!t]
	\centering
	\includegraphics[scale=0.3]{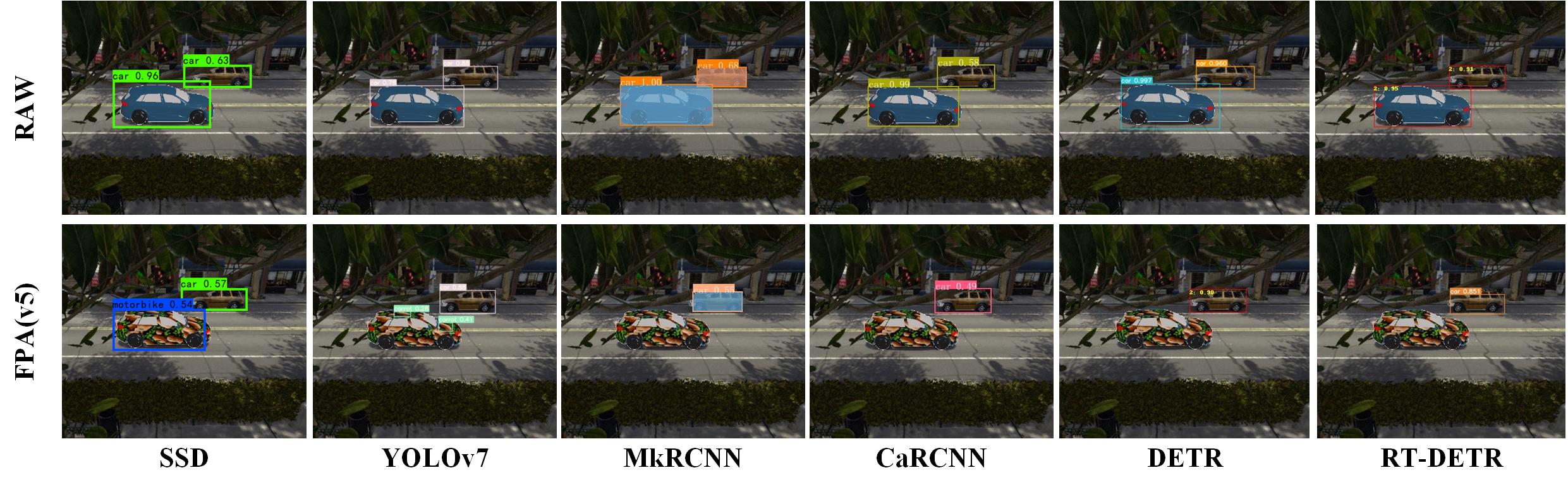}
	\caption{The first row shows the prediction results of original samples under various detectors, and the second row shows the prediction results of our adversarial camouflage, FPA(v5).}
	\label{fig:3}
\end{figure*}

\noindent{\textbf{Transferability Evaluation Across Different Architecture Detectors }}
To assess the effectiveness of adversarial camouflage on unknown models, we introduced several different architectures of detection models, including SSD, YOLOv7, Mask R-CNN(MkRCNN), Cascade R-CNN(CaRCNN), DETR, and RT-DETR. The test results for AP@0.5 are shown in Table.\ref{lab:3}, where our adversarial camouflage, FPA(v5), outperforms others on every detection model. In the latest detection model YOLOv7, it achieves a significant reduction from the original 97.9\% to 11.2\%, demonstrating robust attack effectiveness. When confronted with CaRCNN, while other camouflage methods only slightly decrease AP@0.5, our adversarial camouflage, FPA(v5), achieves a reduction of 48.2\%. Interestingly, when facing detection models based on Transformer architecture, our attack method also demonstrates strong attack capability, achieving a reduction of 68.3\% in DETR. This indicates the robustness of our method against unknown architectures of black-box models. From the table, we can see that the overall attack effect of RAUCA is slightly lower than that of FPA. However, RAUCA is trained based on a constructed weather dataset (40,960 images). In order to achieve anti-interference of environmental factors, an environmental feature extractor is added to combine with Neural Render. Compared with the method in this paper, the structure is more complex, the amount of data required is larger, the deployment is more cumbersome, and the attack effect is slightly worse. Figure.\ref{fig:3} illustrates the prediction results of our adversarial camouflage, FPA(v5), across different detection models. Our method enables vehicles to evade detection model recognition, further validating the robust transferability of our approach.

\begin{table*}[!ht]
	\caption{The AP@0.5 test results of adversarial camouflage under different detection models.}
	\centering
        \small                
	\begin{tabular}{c|c|c|c|c|c|c}
		\hline
		Methods & \multicolumn{2}{c|}{Single-Stage} & \multicolumn{2}{c|}{Two-Stage} & \multicolumn{2}{c}{Transformer} \\ \hline
		~ & SSD & YOLOv7 & MkRCNN & CaRCNN & DETR & RT-DETR \\ \hline
		RAW & 0.846 & 0.979 & 0.961 & 0.988 & 0.969 & 0.798 \\ \hline
		Noise & 0.716 & 0.874 & 0.917 & 0.968 & 0.719 & 0.426 \\ \hline
		CAMOU & 0.447 & 0.831 & 0.849 & 0.967 & 0.854 & 0.251 \\ \hline
		ER & 0.503 & 0.896 & 0.899 & 0.978 & 0.866 & 0.309 \\ \hline
		DAS & 0.618 & 0.763 & 0.633 & 0.885 & 0.745 & 0.672 \\ \hline
		FCA & 0.205 & 0.259 & 0.597 & 0.872 & 0.389 & 0.257 \\ \hline
		DTA & 0.436 & 0.808 & 0.867 & 0.952 & 0.729 & 0.261 \\ \hline
		ACTIVE & 0.247 & 0.492 & 0.608 & 0.787 & 0.629 & 0.297 \\ \hline
		RAUCA & 0.214 & 0.313 & 0.559 & 0.772 & 0.382 & 0.209 \\ \hline
		FPA(bk) & 0.174 & 0.098 & 0.493 & 0.557 & 0.327 & 0.067 \\ \hline
  		FPA(v3) & 0.197 & 0.358 & 0.476 & 0.756 & \textbf{0.138} & 0.081 \\ \hline
		FPA(v5) & 0.179 & \textbf{0.112} & \textbf{0.406} & \textbf{0.506} & 0.286 & \textbf{0.053} \\ \hline
		FPA(FR) & \textbf{0.117} & 0.481 & 0.616 & 0.847 & 0.400 & 0.200 \\ \hline
	\end{tabular}
	\label{lab:3}
\end{table*}

\begin{figure}[!t]
	\centering
	\includegraphics[width=1\linewidth]{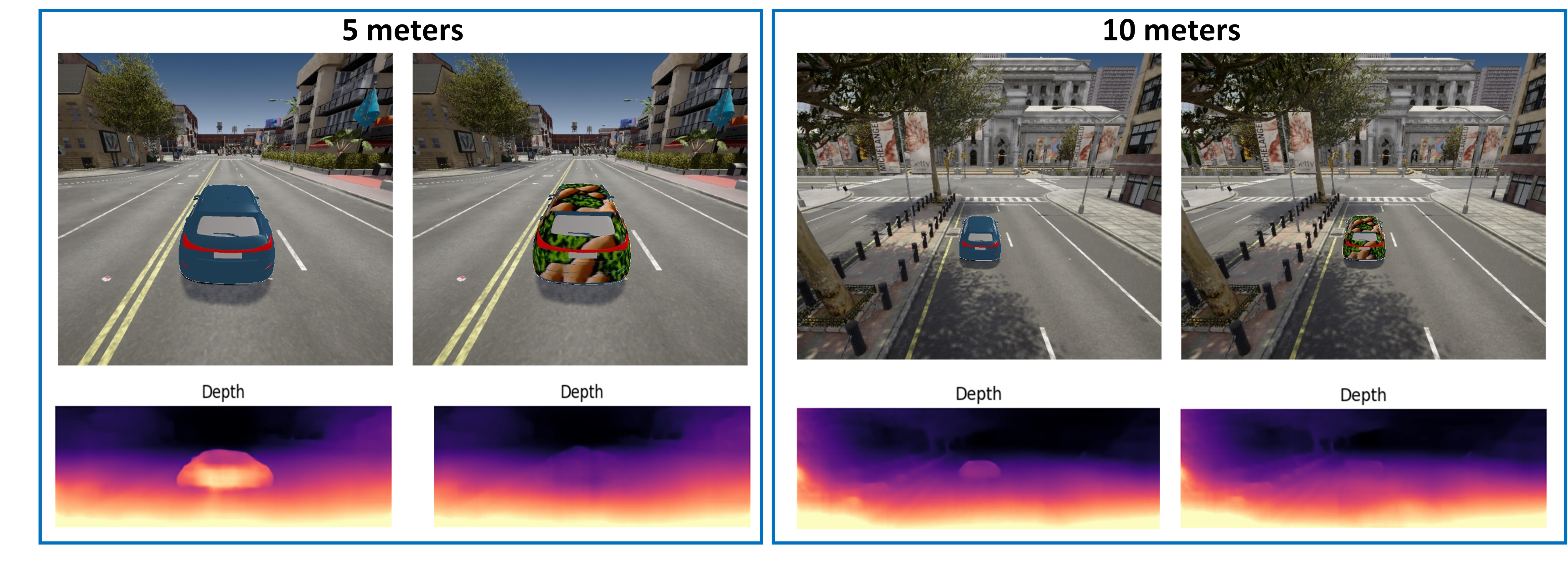}
	\caption{The prediction results of the original vehicle and the vehicle covered with our adversarial camouflage under the depth estimation model.}
	\label{fig:4}
\end{figure}

\noindent{\textbf{Transferability Evaluation Across Different Tasks }}  We aim for the adversarial camouflage, once covering the vehicle, to not only impact target detection recognition but also disrupt other tasks, such as depth estimation commonly used in the field of autonomous driving. We validate the attack performance of the adversarial camouflage on publicly trained depth estimation models, including Monodepth2 (MoD for short)\cite{godard2019digging}, Depth Hints (DH for short)\cite{watson2019self}, and ManyDepth (MaD for short)\cite{watson2021temporal}. For a more comprehensive evaluation, we generated 24 test images based on different environments at distances of 5 meters and 10 meters. Additionally, we selected two common metrics on depth estimation, depth error$({\mathcal{E}}_d)$, and the proportion of affected regions$(\mathcal{R}_{a})$\cite{cheng2022physical}. The experimental results, as presented in Table.\ref{lab:4}, indicate that our attack achieves the best performance on MoD, with a depth influence of 12.36 and a proportion of affected regions reaching 85.4\%. Figure.\ref{fig:4} illustrates the prediction results on MoD. After applying the adversarial camouflage, the display of the original vehicle depth almost completely disappears, further confirming the strong transferability of our method across different tasks.

\begin{table}[!ht]
	\caption{Results of depth error and the proportion of affected regions testing on trained depth estimation models after applying camouflage coverage.}
	\centering
        \small
	\begin{tabular}{c|c|c|c}
		\hline
		Evaluation Metrics & MoD & DH & MaD \\ \hline
		${\mathcal{E}}_d$ & 12.36 & 7.07 & 7.88 \\ \hline
		$\mathcal{R}_{a}$ & 85.40 & 73.93 & 91.76 \\ \hline
	\end{tabular}
	\label{lab:4}
\end{table}

\subsection{Simulation Evaluation of Physical World Factors}
\begin{figure}[!t]
	\centering
	\includegraphics[width=1\linewidth]{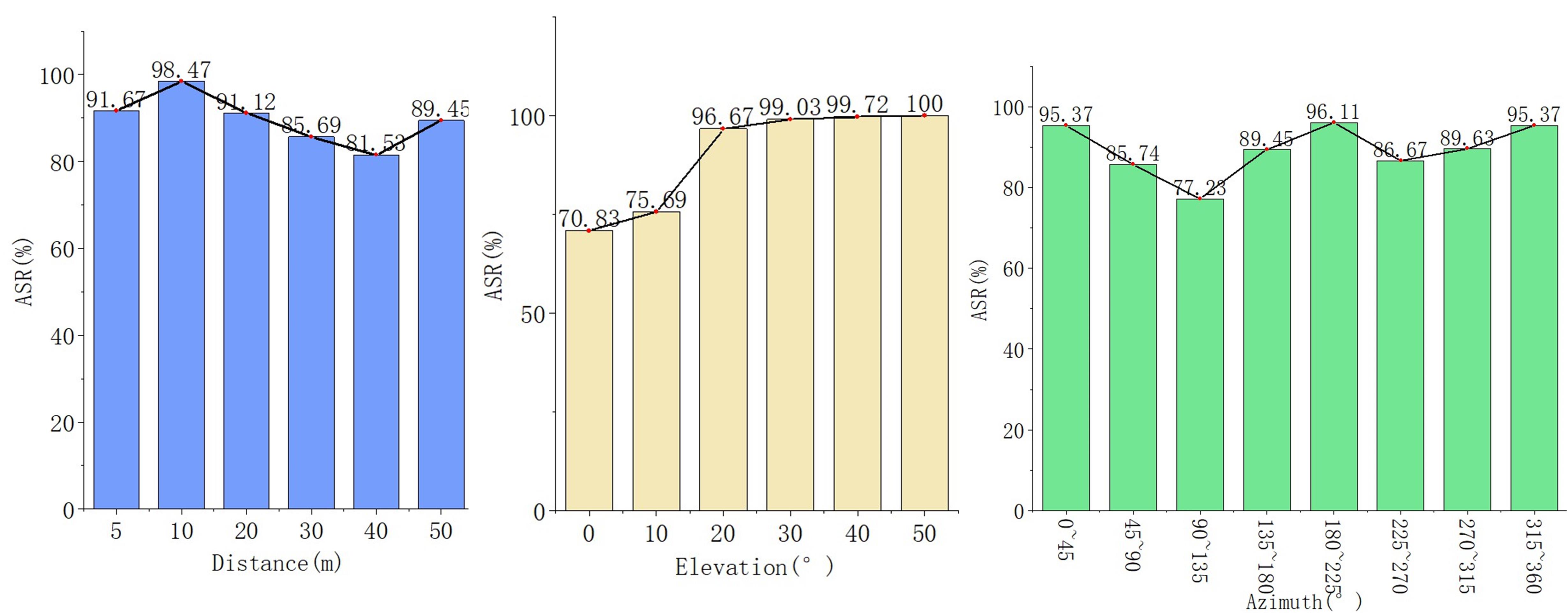}
	\caption{The ASR test results under different distances, pitch angles, and azimuth angles.}
	\label{fig:5}
\end{figure}

\noindent{\textbf{Robustness Evaluation at Different Locations }}  In the physical world, different distances and angles can significantly affect the effectiveness of adversarial camouflage. To validate the robustness of our adversarial camouflage to positional changes, we utilized images generated using PyTorch3D rendering (pure background) for evaluation. Specifically, most previous studies did not conduct validation at longer distances. To comprehensively evaluate the adversarial camouflage, the distances of the images we collected include [5, 10, 20, 30, 40, 50]. We captured images every 3 degrees within a 360° range, with a fixed range of pitch angles [0, 10, 20, 30, 40, 50], resulting in a total of 4320 images. Validation results are shown in Figure.\ref{fig:5}, evaluated based on distance, pitch angle, and azimuth angle. To better represent the validation results, azimuth angles are evaluated within a range of 45° increments. Note that as the distance increases, the recognition capability of the target detection model is affected beyond 40 meters. We found that at 40 meters, the ASR of the detection model on the original vehicle is 2.92\%, while at 50 meters, it is 15.14\%. Meanwhile, we noticed a slight decrease in ASR after 10 meters, followed by an increase after 40 meters, with overall fluctuations within a small range as the distance increases. Due to the partial coverage of our camouflage, which does not completely conceal parts of the vehicle such as tires, windows, and grilles, images captured from certain angles with more of these features may result in a slight decrease in ASR. Nonetheless, under angle variations, the lowest ASR can reach up to 70.83\%, indicating that our method exhibits good robustness across different positions.

\begin{figure}[!t]
	\centering
	\includegraphics[scale=0.22]{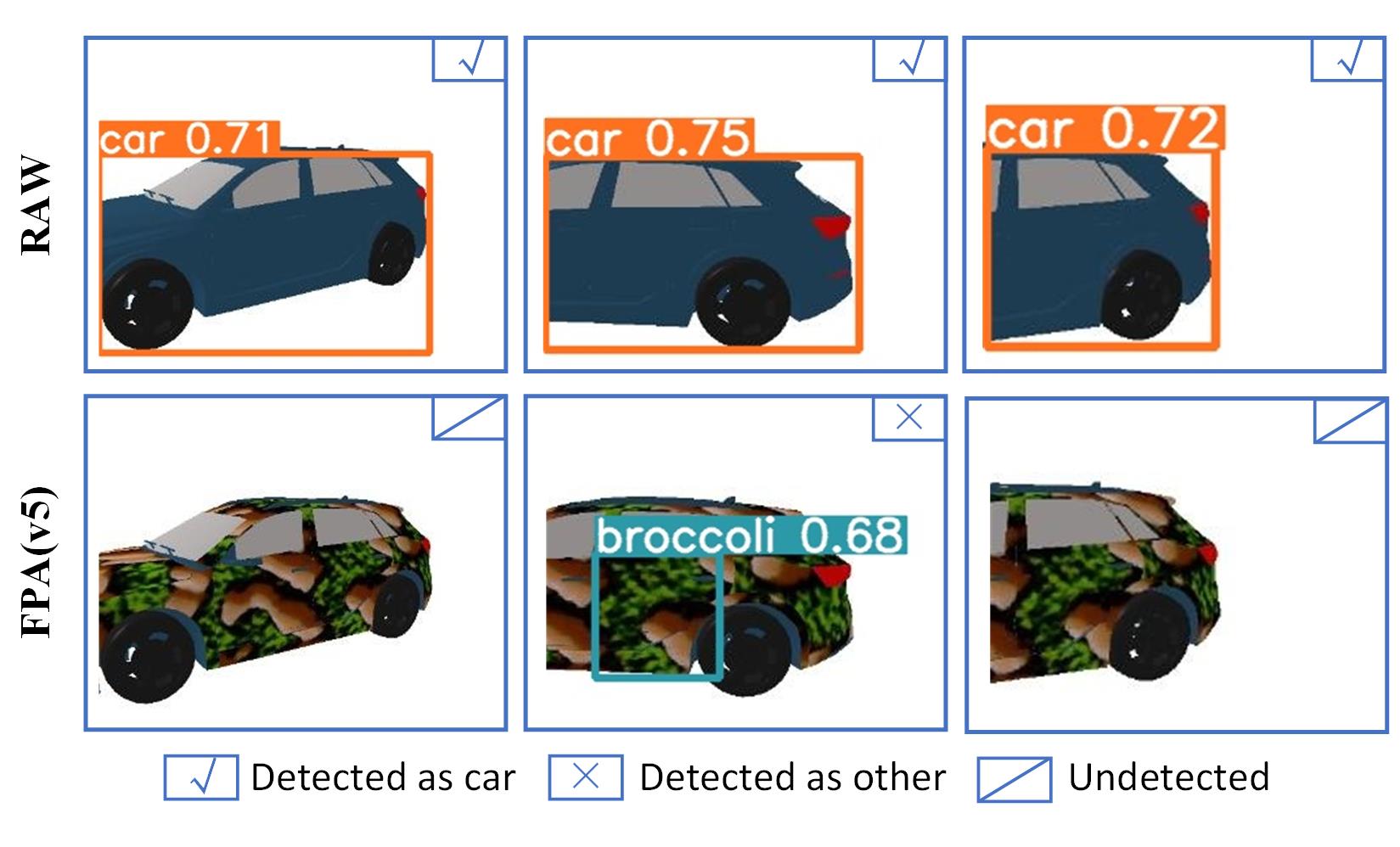}
	\caption{The prediction results of the original car and the camouflaged car under YOLOv5. The ticked box indicates a correct recognition result, the crossed box indicates a recognition error, and the slashed box indicates no recognition result.}
	\label{fig:6}
\end{figure}

\noindent{\textbf{Robustness Evaluation under Different Occlusions }}  In the physical world, targets often appear partially visible in images, leading to partial camouflage coverage. Therefore, we conducted experiments to evaluate the robustness of our method under different degrees of occlusion. To prevent images from becoming too small after occlusion, we set the distance range to [5, 10, 15, 20], pitch angles to [15, 30, 45, 60], and collected images at intervals of 12° within a 360° angle range. To generate rendered images with different occlusion levels, we start with the width of the target as the initial length and occlude it from left to right in 10 equal proportions from 0 to 1. Meanwhile, we designate proportions from 0.1 to 0.3 as small occlusion, proportions from 0.4 to 0.6 as middle occlusion, and proportions from 0.7 to 0.9 as large occlusion. Hence, each data point in Table.\ref{lab:5} represents the ASR calculated across 360 test images. In Figure.\ref{fig:6}, we present the prediction results under occlusion ratios of [0.1, 0.4, 0.7]. From the data in the table, it is evident that our camouflage achieves an ASR close to 100\% under different occlusion areas, indicating the strong robustness of our method against occlusion.
\begin{table*}[!ht]
	\caption{The ASR test results of the adversarial camouflage under different occlusion ranges.}
	\centering
        \small
	\begin{tabular}{c|c|c|c|c|c|c|c|c}
		\hline
		\multirow{3}{*}{
			\makecell{Occlusion \\ range}} & \multicolumn{8}{c}{Distance} \\ \cline{2-9}
		& \multicolumn{2}{c|}{5} & \multicolumn{2}{c|}{10} & \multicolumn{2}{c|}{15} & \multicolumn{2}{c}{20} \\ \cline{2-9}
		& RAW & FPA(v5) & RAW & FPA(v5) & RAW & FPA(v5) & RAW & FPA(v5) \\ \hline
		None & 1.67 & 100 & 0.00 & 100 & 0.00 & 100 & 0.00 & 98.33 \\ \hline
		Small & 18.06 & 100 & 3.05 & 100 & 4.17 & 100 & 7.23 & 99.45 \\ \hline
		Middle & 44.45 & 100 & 30.39 & 100 & 39.67 & 100 & 56.67 & 100 \\ \hline
		Large & 86.12 & 100 & 82.23 & 100 & 84.34 & 100 & 89.37 & 100 \\ \hline
	\end{tabular}
	\label{lab:5}
\end{table*}

\begin{figure}
	\centering
	\includegraphics[scale=0.23]{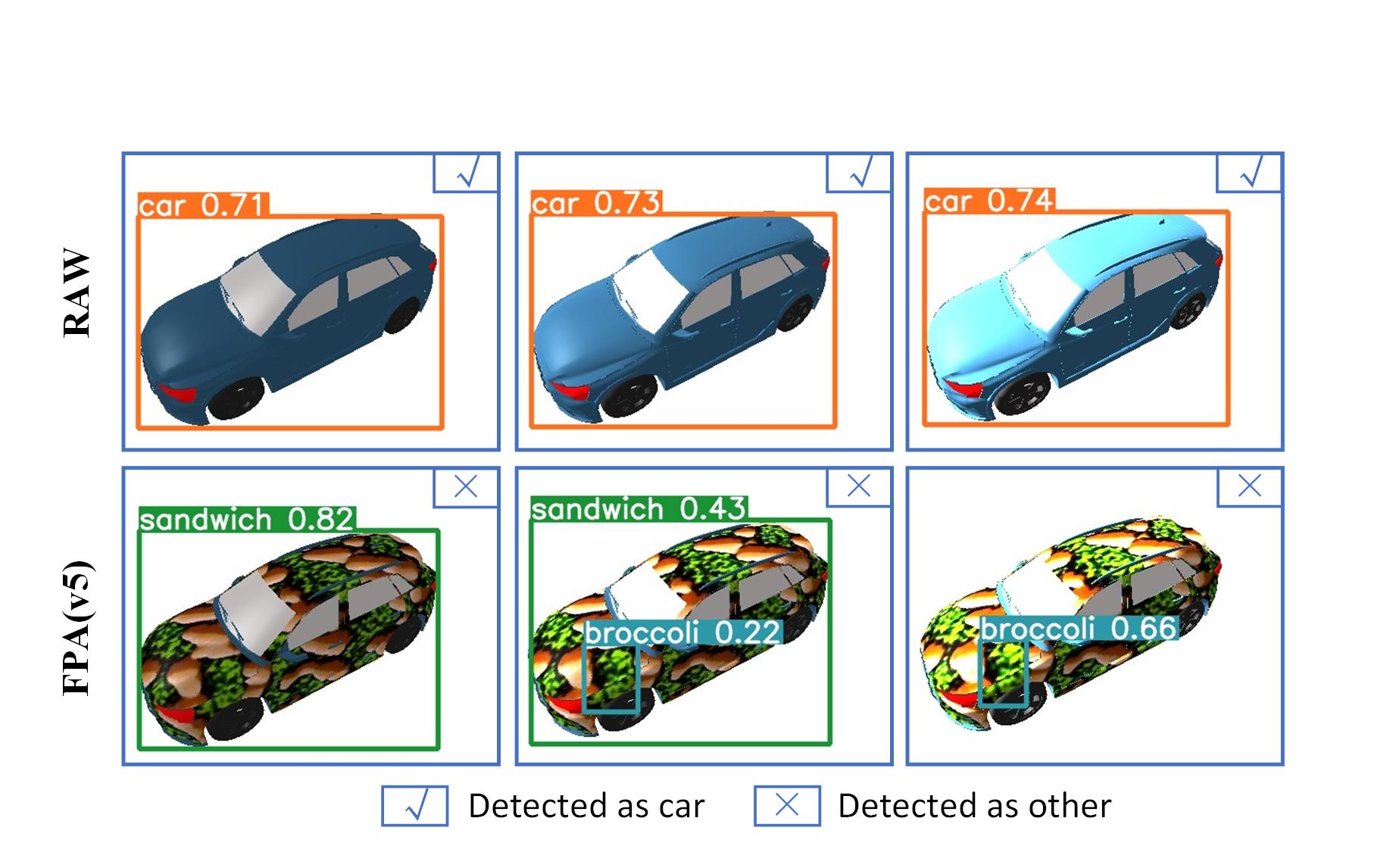}
	\caption{The rendered images with illumination intensities of 0.5, 2.0, and 5.0 (from left to right) and the corresponding prediction results under YOLOv5 are shown.}
	\label{fig:7}
\end{figure}

\noindent{\textbf{Robustness Evaluation under Different Reflection Coefficients }}  The intensity of lighting in the physical world can significantly affect adversarial camouflage. This is primarily because varying illumination intensities can result in different degrees of reflection on the surface of the vehicle, consequently altering the camouflage.  Our method simulates physical parameters during the training process of adversarial camouflage using PyTorch3D, ensuring robustness in real-world environments. 
Therefore, in this part of the experiment, we used PyTorch3D to set the environmental illumination intensity ($\upsilon_{light}$) to simulate the changes in the target surface under different conditions. Specifically, we defined a set of illumination intensity coefficients [0.5, 1.0, 2.0, 3.0, 5.0] for the experiments. The data collection method was the same as in the occlusion experiments. Each value in Table \ref{lab:6} is calculated based on the ASR from 480 test images. From the data in the table, it can be seen that the ASR of the adversarial camouflage fluctuates around 97\%, showing little to no impact from changes in illumination. The graphs depicted in Figure \ref{fig:7} illustrate that the camouflaged vehicles successfully evade detection by the target detection system, indicating that our method exhibits strong robustness under varying lighting conditions.

\begin{table}[!ht]
	\caption{The ASR test results of the adversarial camouflage under different illumination intensity settings.}
	\centering
	\begin{tabular}{c|c|c|c|c|c}
		\hline
		diffuse\_color & 0.5 & 1.0 & 2.0 & 3.0 & 5.0 \\ \hline
		RAW & 1.67 & 1.67 & 1.25 & 0.00 & 0.83 \\ \hline
		FPA(v5) & 97.91 & 97.71 & 97.08 & 96.25 & 96.87 \\ \hline
	\end{tabular}
	\label{lab:6}
\end{table}

\subsection{Concealment Evaluation}
The purpose of camouflage is to reduce the optical contrast between the target and the background, thereby increasing the similarity between the target and the background, and thus achieving the concealment of the target. Therefore, in adversarial camouflage, we not only need to effectively attack the target detection model but also ensure that the camouflage maintains a level of similarity with the background, enhancing visual concealment. Most previous work on the concealment of adversarial camouflage\cite{hu2021naturalistic,lin2023diffusion} relied on subjective observational experiments to measure the effectiveness of target camouflage, lacking scientific and accurate assessment. We conducted experiments based on digital image analysis techniques to comprehensively and accurately analyze the effectiveness of camouflage, focusing on model attention and image similarity.

\begin{figure}
	\centering
	\includegraphics[width=1.0\linewidth]{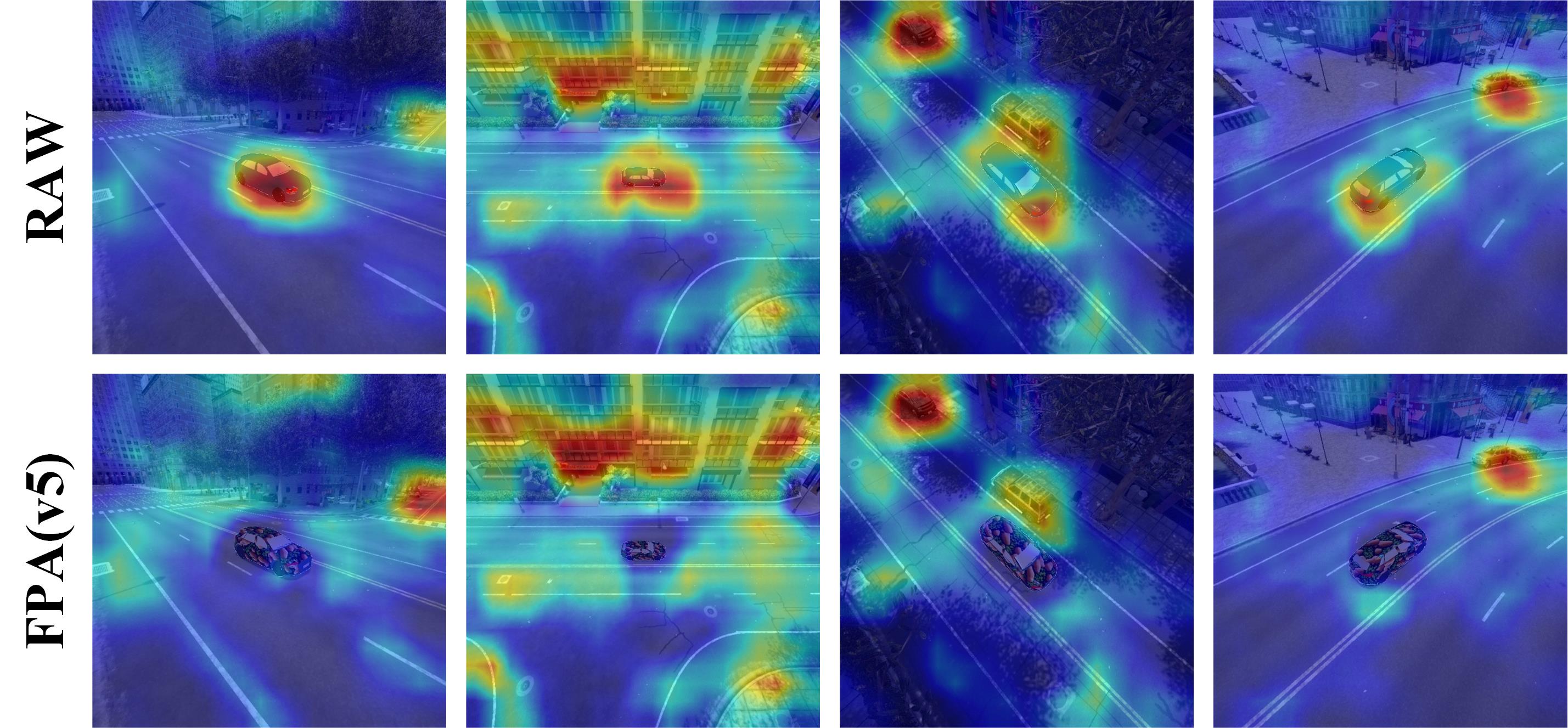}
	\caption{The attention distribution results generated by Grad-CAM for the original and camouflaged vehicles.}
	\label{fig:8}
\end{figure}

\noindent{\textbf{Evaluation of Model Attention }}  We analyze the interference of adversarial camouflage on model attention using the Grad-CAM visualization tool. Grad-CAM utilizes a pre-trained ResNet50 as the base model for recognizing images. The findings depicted in Figure.\ref{fig:8} reveal a noteworthy pattern: the model's attention predominantly centers around the original vehicle, with minimal focus directed towards the camouflaged vehicle. This observation underscores the disruptive effect of adversarial camouflage, effectively diffusing the model's attention away from the intended target.

\begin{figure}[thp]
	\centering
	\includegraphics[width=1.0\linewidth]{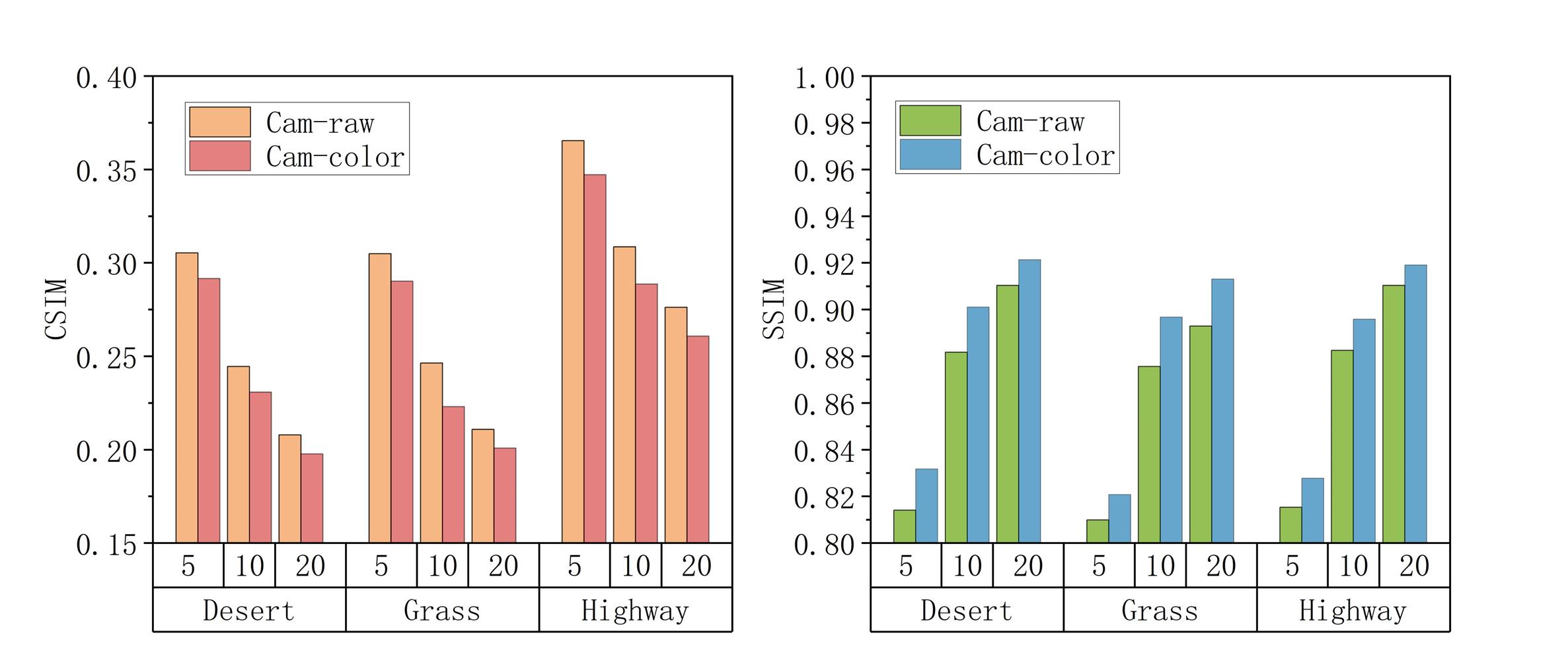}
	\caption{The comparison of similarity between adversarial camouflages generated under two training modes in desert, grassland, and highway scenarios.}
	\label{fig:9}
\end{figure}

\begin{figure}
	\centering
	\includegraphics[width=0.8\linewidth]{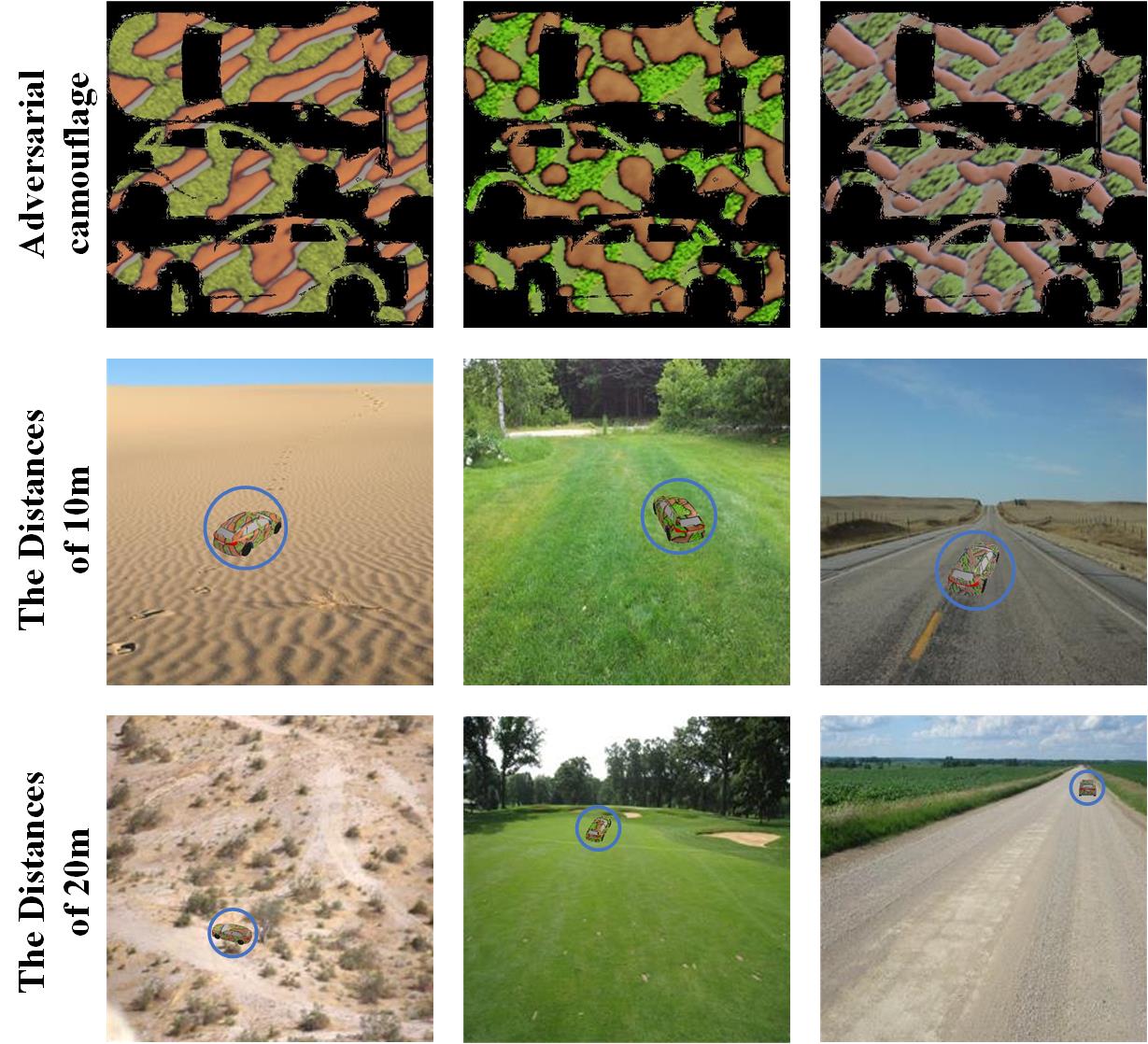}
	\caption{The first row illustrates the adversarial camouflage across various scenarios obtained through Camou\_color training. The second and third rows display the rendered results of the camouflaged car at distances of 10m and 20m, respectively.}
	\label{fig:10}
\end{figure}

\noindent{\textbf{Evaluation of Image Similarity }}  In the constructed scene dataset, we adopt two training modes: Camou-raw, which does not include the $L_{CR}$, and Camou-color, which includes the $L_{CR}$. We generated adversarial camouflage for three different scenarios. To analyze the camouflage effectiveness more scientifically and accurately, we propose two metrics based on image similarity: Color Similarity (CSIM) and Structural Similarity (SSIM)\cite{wang2004image}. Specifically, we first compute the color histograms of the two images. Then, we calculate the histogram difference using the Bhattacharyya distance\cite{choi2003feature} to obtain the color similarity. A smaller value indicates that the colors of the two images are more similar. We compute the similarity between the target and the background within ranges of 5m, 10m, and 20m, and validate it in the constructed dataset. The results are shown in Figure.\ref{fig:9}. 
In all three scenarios, the adversarial camouflage generated by Camou-color achieved better results in both SSIM and CSIM. Table.\ref{lab:7} presents the ASR results of adversarial camouflage under different training modes in YOLOv5, where ``None" indicates the detection results of the vehicle without camouflage. The attack effectiveness of adversarial camouflage under the Camou-color training mode is slightly lower compared to Camou-raw, but both can significantly impact the target detection model. Figure.\ref{fig:10} illustrates rendered images of vehicles concealed with camouflage across different scenarios. Visually, these images create confusion, indicating that our method significantly enhances camouflage concealment while maintaining attack effectiveness.

\begin{table}[!ht]
	\caption{The original vehicle, along with the ASR test outcomes of the adversarial camouflage in various scenarios obtained through the two training methodologies.}
	\centering
	\begin{tabular}{c|c|c|c}
		\hline
		Method & Desert & Grass & Highway \\ \hline
		None & 0.50 & 0.35 & 0.77 \\ \hline
		Camou-raw & 91.08 & 92.59 & 87.86 \\ \hline
		Camou-color & 83.56 & 87.27 & 84.89 \\ \hline
	\end{tabular}
	\label{lab:7}
\end{table}

\subsection{Physical World Evaluation}
\begin{figure}
	\centering
	\includegraphics[width=1.0\linewidth]{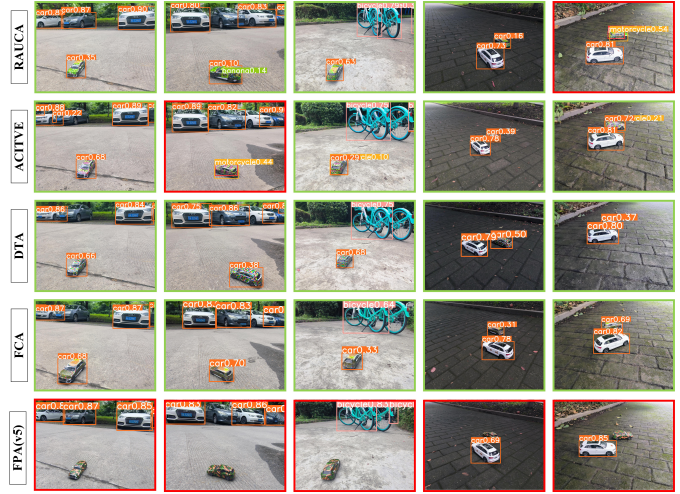}
	\caption{The prediction results of the original model car and the camouflaged cars in the physical world. Green boxes indicate correct detections by the detector, while red boxes indicate successful attacks by the adversarial camouflage.}
	\label{fig:11}
\end{figure}

In this section, we evaluated the adversarial camouflage's performance in the physical world. Specifically, we conducted experiments in the physical world using a 1:24 scale Audi Q5 car model. Due to budget and time constraints, we utilized an HP color printer to print our adversarial camouflage FPA(v5), along with four comparative camouflages: RAUCA, ACTIVE, DTA, and FCA. We then cut out the camouflaged sections and adhered them to the corresponding areas of the car models. We captured surround shots of the target at three different distances within three distinct environments to assess the efficacy of camouflage, utilizing the Redmi K50 Pro as the imaging device. The original and camouflaged cars were captured using the same model for data collection. We collected 100 images for each type of camouflage to validate the results. The AP@0.5 results are shown in Table.\ref{lab:8}, and some of the prediction results can be seen in Figure.\ref{fig:11}. Our adversarial camouflage significantly reduces AP@0.5 across different detection models. Notably, it achieves a reduction of up to 66.7\% in the SSD model. While other adversarial camouflages struggle to generate effective attacks against the two-stage detector Faster R-CNN, our camouflage successfully lowers AP@0.5 to 0.529. This demonstrates that our method has stronger attack effects in the physical world than previous work.

\begin{table}[!ht]
	\caption{The AP@0.5 test results of the original car and the camouflaged car in the physical world.}
	\centering
	\begin{tabular}{c|c|c|c|c}
		\hline
		Method & SSD & Yolov5 & FRCNN & DETR  \\ \hline
		RAW & 0.930 & 0.995 & 0.995 & 0.985 \\ \hline
		RAUCA & 0.395 & 0.586 & 0.758 & 0.406 \\ \hline
		ACTIVE & 0.571 & 0.675 & 0.891 & 0.607 \\ \hline
		DTA & 0.854 & 0.875 & 0.912 & 0.725 \\ \hline
		FCA & 0.713 & 0.862 & 0.874 & 0.621 \\ \hline
		FPA & \textbf{0.263} & \textbf{0.442} & \textbf{0.529} & \textbf{0.315} \\ \hline
	\end{tabular}
	\label{lab:8}
\end{table}

\subsection{Ablation Studies}
\label{sec:abl}
In this section of the experiment, we investigated the effects of different training methods (including diffusion model modules and initialization areas) and terms of loss function on adversarial camouflage.

\begin{figure}
	\centering
	\includegraphics[width=1.0 \linewidth]{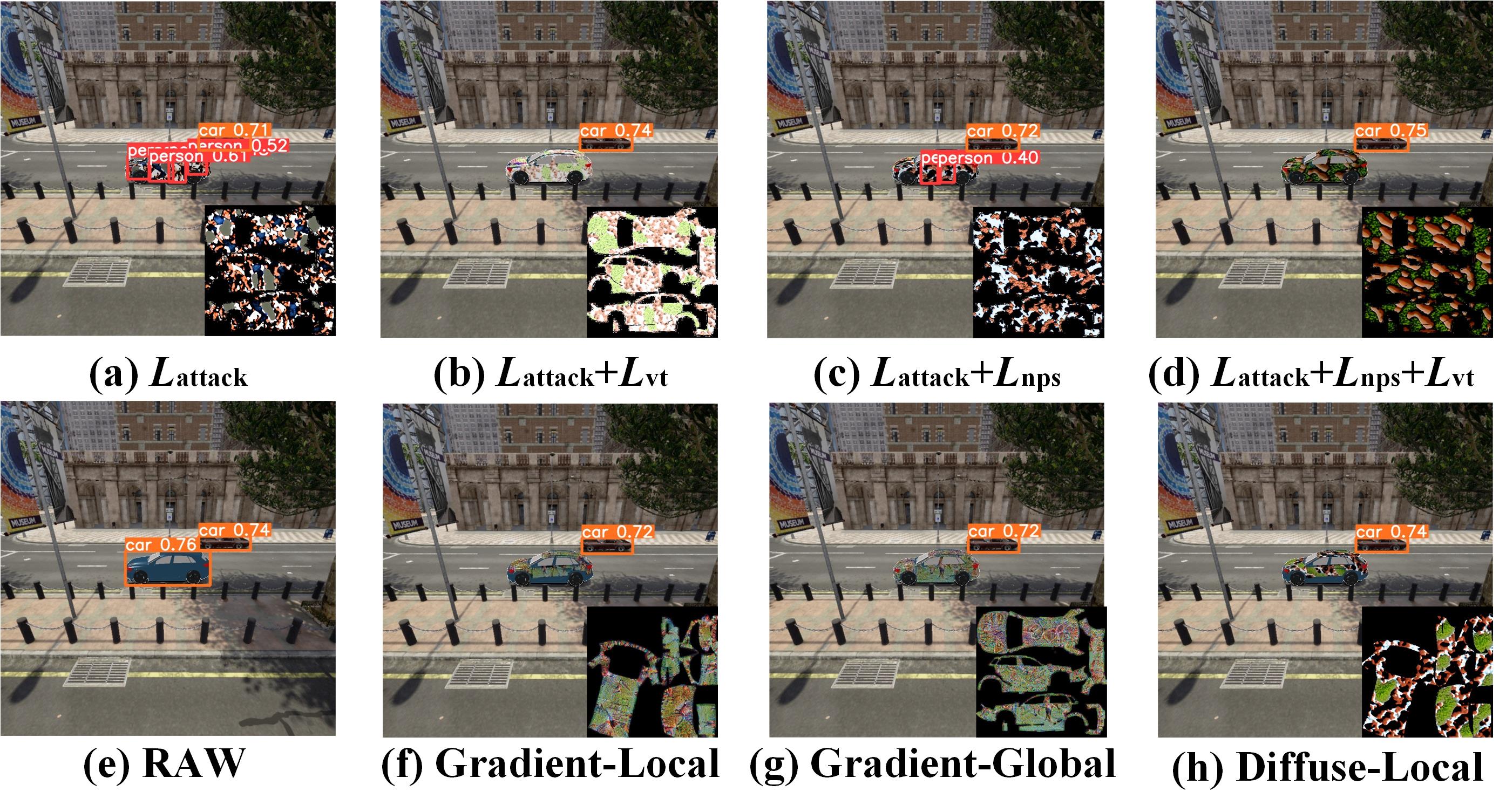}
	\caption{Confrontation camouflage display and prediction results in different settings.}
	\label{fig:12}
\end{figure}

\noindent{\textbf{Different Combinations of Loss Functions. }}  We investigated the impact of different combinations of loss functions on adversarial camouflage, including ${L_{attack}}$, ${L_{attack}+L_{vt}}$, and ${L_{attack}+L_{nps}}$ for validation. Meanwhile, we compared the AP@0.5 computed on the test set collected based on Carla. The experimental results are presented in Table.\ref{lab:9}, indicating a decrease in the performance of adversarial camouflage in the absence of certain terms of the loss function. When considering only the ${L_{attack}}$ term, the attack effectiveness surpasses the other two experimental groups. However, as depicted in Figure.\ref{fig:12}, the absence of the ${L_{vt}}$ term leads to numerous black spots in the camouflage, resulting in reduced concealment. In contrast, the lack of the constraint ${L_{nps}}$ results in brighter camouflage colors, leading to significant loss of information of the pixels during printing. Therefore, combining ${L_{attack}}$, ${L_{nps}}$, and ${L_{vt}}$, we achieve stronger attack performance and better concealment for adversarial camouflage.
\begin{table}[!ht]
	\caption{The comparison results of AP@0.5 under different loss function combination schemes.}
	\centering
 \small
	\begin{tabular}{m{0.8in}|c|c|p{0.4in}|c}
		\hline
		Method  & YOLOv3 & YOLOv5 & FRCNN & DETR \\ \hline
		${L_{attack}}$ & 0.286 & 0.226 & 0.442 & 0.284 \\ \hline
		${L_{attack}}+L_{vt}$ & 0.626 & 0.534 & 0.612 & 0.780 \\ \hline
		${L_{attack}}+L_{nps}$ & 0.506 & 0.281 & 0.468 & 0.416 \\ \hline
		${L_{attack}}+L_{vt}+L_{nps}$ & 0.159 & 0.088 & 0.319 & 0.286 \\ \hline
	\end{tabular}
	\label{lab:9}
\end{table}

\noindent{\textbf{Different Training Methods }}  In this section, we investigate the impact of the diffusion model on optimizing against camouflage and the effect of the initialization area on the attack's effectiveness. We establish two optimization methods: traditional gradient-based optimization and diffusion model-based optimization. Meanwhile, two methods of initial area change are set, local coverage and global coverage. Local coverage refers to partial regions on structures such as car doors and roofs, while global coverage refers to the camouflage covering the entire vehicle (excluding structures such as windows and tires). Specifically, based on the training data collected from Carla and using YOLOv5 as the victim detection model, we generated four sets of adversarial camouflages: 
\begin{itemize}
    \item Gradient-Local: Adversarial camouflage generated based on gradient optimization and local coverage.
    \item Gradient-Global: Adversarial camouflage generated based on gradient optimization and global coverage.
    \item Diffuse-Local: Adversarial camouflage generated based on diffusion model optimization and local coverage.
    \item Diffuse-Global: Adversarial camouflage generated based on diffusion model optimization and global coverage.
\end{itemize}

Where Diffuse-Global represents the FPA(v5) adversarial camouflage used in the previous comparison. The AP@0.5 results of the adversarial camouflage under different detection models are presented in Table.\ref{lab:10}. It is observed that the diffusion model optimization approach significantly outperforms the gradient approach. Meanwhile, enlarging the coverage area significantly enhances the effectiveness of camouflage. As observed in Figure.\ref{fig:12}, both gradient and diffusion model optimization-based camouflages effectively conceal the car, underscoring the robustness of our training methodology and the design of loss functions.
\begin{table}[!ht]
	\caption{Comparison of the AP@0.5 results of adversarial camouflages generated by different optimization methods and initial areas under different detection models.}
	\centering
       \small
	\begin{tabular}{c|c|c|p{0.4in}|c}
		\hline
		Method & YOLOv3 & YOLOv5 & FRCNN & DETR \\ \hline
		Gradient-Local & 0.676 & 0.565 & 0.681 & 0.785 \\ \hline
		Gradient-Global & 0.527 & 0.493 & 0.605 & 0.664 \\ \hline
		Diffuse-Local & 0.481 & 0.415 & 0.536 & 0.436 \\ \hline
		Diffuse-Global & 0.159 & 0.088 & 0.319 & 0.286 \\ \hline
	\end{tabular}
	\label{lab:10}
\end{table}

\section{Discussions}
\noindent{\textbf{Implications }} In conclusion, our work presents a novel approach to adversarial camouflage through the integration of a comprehensive 3D rendering framework, departing from conventional neural rendering methods. Using the inherent capabilities of the framework, we achieved a nuanced representation of 3D targets by simulating authentic lighting conditions and material variations, thereby contributing to a more realistic rendering outcome. The combination of a generatively learned adversarial pattern from diffusion models, along with a designed adversarial loss and covert constraint loss, ensures the adversarial and covert nature of the camouflage in the physical world. The contribution of our work lies in the establishment of a rendering system rooted in a robust 3D framework, which enhances rendering efficiency and enables the generation of adversarial camouflage in sticker mode for practical deployment. The incorporation of confrontation loss and concealment constraint loss further strengthens the aggressiveness and concealment of the camouflage, demonstrating adaptability across diverse environments. In future works, we aim to explore additional refinements to adversarial patterns, investigate real-world deployment scenarios, and evaluate the generalization of our approach to different objects and environments, thereby advancing the field of adversarial camouflage in both theoretical and practical dimensions.

\noindent{\textbf{Limitations }} 
Due to the limitations of the renderer's capabilities, FPA is unable to simulate deformation during the rendering process. Consequently, when dealing with non-rigid targets, the effectiveness of the generated adversarial camouflage in real-world physical tests may exhibit significant deviations. In the future, we will further investigate adversarial texture generation techniques for 3D non-rigid targets.

\section*{Acknowledgment}
This work is funded by the National Natural Science Foundation of China (No.62103330, 62233014).



\bibliographystyle{plain}
\bibliography{sample}

\begin{IEEEbiographynophoto}
{Yang Li}
is an associate professor with the school of automation at Northwestern Polytechnical University, Xi’an, China. After receiving his bachelor's and doctoral degrees from Northwestern Polytechnical University in 2014 and 2018 respectively, he worked as a research fellow in SenticTeam under Professor Erik Cambria at Nanyang Technological University in Singapore and also was an adjunct research fellow at the A*STAR High-Performance Computing Institute (IHPC). His research goal is to build a trustworthy AI system in real application, and his research interests are in natural language processing, machine learning, recommendation systems, explainable artificial intelligence, etc. He has published several papers on these topics at international conferences and in peer-reviewed journals. He is an active reviewer of several journals, e.g., INFORM FUSION, IEEE TAFF, NEUCOM, KBS, KAIS, etc. He also is a guest editor of Future Generation Computer Systems.  
\end{IEEEbiographynophoto}

\begin{IEEEbiographynophoto}
{Wenyi Tan} received the bachelor's degree from the School of Automation, Wuhan University of Technology, China, in 2022. He is working toward the master's degree from  the School of Automation, Northwestern Polytechnical University. His research interests are in Adversarial Attack \& Defense in AI, Explainable Artificial Intelligence, Object Detection, Adversarial Camouflage Generation, etc.
\end{IEEEbiographynophoto}

\begin{IEEEbiographynophoto}
{Tingrui Wang} received her bachelor's degree from the Civil Aviation Flight University of China, in 2023. She is currently working toward the master's degree with the College of Automation, Northwestern Polytechnical University, Xi'an, China.Her research interests include adversarial attack and deep learning.
\end{IEEEbiographynophoto}

\begin{IEEEbiographynophoto}
{Xinkai Liang} is pursuing his Ph.D. under Dr.Bin Zhou in the department of Astronautics, Harbin Institute of Technology. Currently he is working as Senior Engineer at Science and Technology on Complex System Control and Intelligent Agent Cooperation Laboratory, China. He got his BS degree from Zhengzhou University in 2014 and MS degree from Beihang University in 2017. His research interests include deep learning, image processing and SLAM.
\end{IEEEbiographynophoto}

\begin{IEEEbiographynophoto}
{Quan Pan}
Born in China in 1961. He received a B.S. degree in automatic control from the Huazhong University of Science and Technology, in 1982, and an M.S. and Ph.D. degrees in control theory and application from Northwestern Polytechnical University. From 1991 to 1993, he was an Associate Professor at Northwestern Polytechnical University, where he has been a Professor with the Automatic Control Department, since 1997. He has authored 11 books and more than 400 articles. His research interests include information fusion, target tracking and recognition, deep network and machine learning, UAV detection navigation, and security control, polarization spectral imaging and image processing, industrial control system information security, commercial password applications, and modern security technologies. He is an Associate Editor of the journal Information Fusion and Modern Weapons Testing Technology.
\end{IEEEbiographynophoto}

\end{document}